\pdfoutput=1

\documentclass[11]{article}
\usepackage[]{acl}
\usepackage{times}
\usepackage{latexsym}

\usepackage[T1]{fontenc}

\usepackage[utf8]{inputenc}
\usepackage{pifont}
\usepackage{soul}
\usepackage{footmisc}
\usepackage{microtype}
\usepackage{tabularx}
\usepackage{xspace}
\usepackage{makecell}
\usepackage{tipa}
\usepackage{graphicx}
\usepackage{booktabs}
\usepackage{multirow}
\usepackage{caption}
\usepackage{amsmath} 
\usepackage{arydshln}
\usepackage{amssymb}  
\usepackage{xcolor}   
\usepackage{graphicx}
\usepackage{subcaption}
\usepackage{marvosym}
\usepackage{float}
\usepackage{footmisc}

\usepackage{hyperref}
\usepackage{tablefootnote}
\usepackage{xcolor}
\definecolor{cream}{RGB}{255, 253, 208}
\definecolor{LightSkyBlue}{RGB}{219, 231, 252}
\definecolor{DarkSeaGreen}{RGB}{143,188,143}
\definecolor{RosyBrown}{RGB}{188,143,143}
\definecolor{Brown}{RGB}{165,42,42}
\title{Perspective Transition of Large Language Models \\for Solving Subjective Tasks}

\author{Xiaolong Wang\textsuperscript{*,1,3}, Yuanchi Zhang\textsuperscript{*,7}, Ziyue Wang\textsuperscript{1},Yuzhuang Xu\textsuperscript{4}, Fuwen Luo\textsuperscript{1}\\
\bf Yile Wang\textsuperscript{\Letter,5}, Peng Li\textsuperscript{\Letter,2}, Yang Liu\textsuperscript{1,2,6} \\
  \textsuperscript{1}Dept. of Comp. Sci. \& Tech., Institute for AI, Tsinghua University, Beijing, China \\
  \textsuperscript{2}Institute for AI Industry Research (AIR), Tsinghua University, Beijing, China \\
 \textsuperscript{3}Jiuquan Satellite Launch Center (JSLC), Gansu, China\\ 
 \textsuperscript{4} Harbin Institute of Technology, Harbin, China\\
  \textsuperscript{5}\fontsize{11pt}{10pt}\selectfont College of Computer Science and Software Engineering, Shenzhen University, Shenzhen, China\\
  \textsuperscript{6} Jiangsu Collaborative Innovation Center for Language Competence, Jiangsu, China\\
  \textsuperscript{7} Tencent Inc, China\\
  \texttt{wangxl22@mails.tsinghua.edu.cn, wangyile@szu.edu.cn}\\ \texttt{lipeng@air.tsinghua.edu.cn, liuyang2011@tsinghua.edu.cn}
  }

\DefineFNsymbols*{1}{*}
\setfnsymbol{1}
\begin{document}
\maketitle



\renewcommand{\thefootnote}{\fnsymbol{footnote}} 
    \footnotetext[1]{Equal contribution. Work done at Tsinghua University. }
\renewcommand{\thefootnote}{\arabic{footnote}}

\DefineFNsymbols*{1}{\Letter}

\setfnsymbol{1}

\renewcommand{\thefootnote}{\fnsymbol{footnote}} 
    \footnotetext[1]{Corresponding authors.}
\renewcommand{\thefootnote}{\arabic{footnote}}




\begin{abstract}
Large language models (LLMs) have revolutionized the field of natural language processing, enabling remarkable progress in various tasks. Different from objective tasks such as commonsense reasoning and arithmetic question-answering, the performance of LLMs on subjective tasks is still limited, where the perspective on the specific problem plays crucial roles for better interpreting the context and giving proper response. For example, in certain scenarios, LLMs may perform better when answering from an expert role perspective, potentially eliciting their relevant domain knowledge. In contrast, in some scenarios, LLMs may provide more accurate responses when answering from a third-person standpoint, enabling a more comprehensive understanding of the problem and potentially mitigating inherent biases. In this paper, we propose Reasoning through Perspective Transition (RPT), a method based on in-context learning that enables LLMs to dynamically select among direct, role, and third-person perspectives for the best way to solve corresponding subjective problem. Through extensive experiments on totally 12 subjective tasks by using both closed-source and open-source LLMs including GPT-4, GPT-3.5, Llama-3, and Qwen-2, our method outperforms widely used single fixed perspective based methods such as chain-of-thought prompting and expert prompting, highlights the intricate ways that LLMs can adapt their perspectives to provide nuanced and contextually appropriate responses for different problems.

\end{abstract}

\section{Introduction}
\begin{figure}[t!]
    \centering
    \includegraphics[scale=0.92]{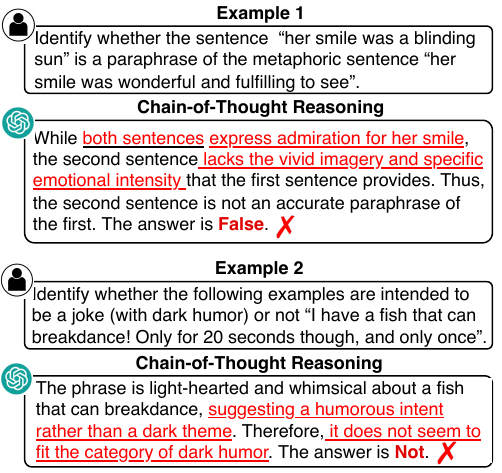}
    \caption{Two examples for showing challenges of solving subjective tasks using chain-of-thought prompting. Refer to Appendix~\ref{sec:more_cases} for more examples.}
    \label{fig:challenge}
    \vspace{-1em}
\end{figure}
\begin{figure*}[t!]
    \centering
    \includegraphics[scale=0.58]{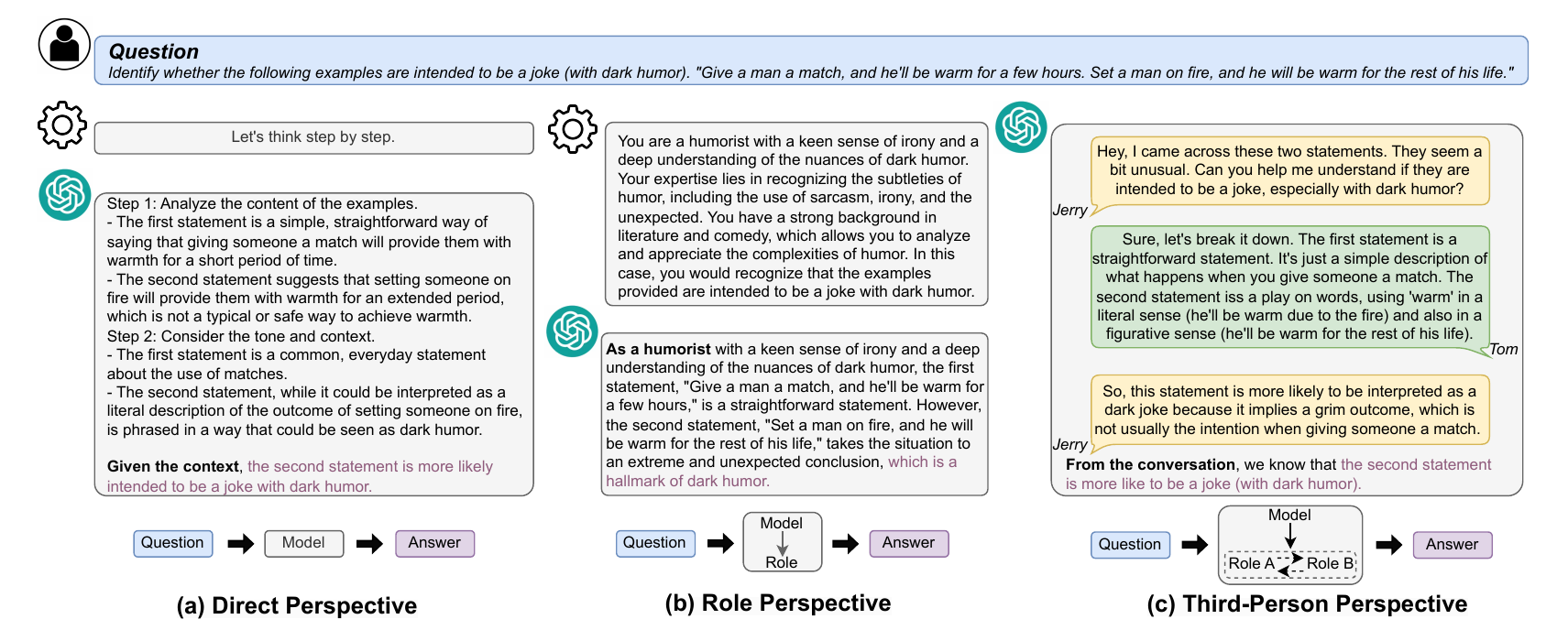}
    \caption{ An example of solving dark humor detection task by different perspectives. (a) direct perspective: the model give the answer according to its analysis ~\cite{kojima2023large}. (b) role perspective: the model gives the answer by setting as a role related to the question ~\cite{xu2023expertprompting}. (c) third-person perspective: the model gives the answer as a third-person based on a simulated dialogue ~\cite{ric}.}
    \label{fig:3perspectives}
    \vspace{-1em}
\end{figure*}
Large language models (LLMs) have exhibited substantial advancements~\cite{gpt3,openaigpt4,openaichatgpt,touvron2023llama,jiang2023mistral} in recent years, demonstrating remarkable performance across various tasks such as mathematical reasoning ~\cite{luo2023wizardmath,yang2023gpt}, code generation ~\cite{chen2021evaluating,roziere2023code}, and commonsense question answering ~\cite{talmor-etal-2019-commonsenseqa}. Meanwhile, research in the realm of \textit{subjective} tasks remains relatively nascent ~\cite{rottger-etal-2022-two,kanclerz-etal-2023-pals,sun2023aligning}. Unlike objective tasks, which are typically well-defined and directly solvable, subjective tasks such as metaphor recognition ~\cite{mohler-etal-2016-introducing} and dark humor detection ~\cite{meaney-etal-2021-semeval} require an understanding of context, linguistic subtleties, and varying individual perspectives. Although advanced chain-of-thought (CoT) style methods have been widely used to elicit the reasoning ability of LLMs ~\cite{cot,kojima2023large,auto-cot,tot}, these approaches primarily focus on ``\textit{how to think deeper in its own perspective}''~\cite{openaio1,deepseek2025r1}, while overlooking ``\textit{from which perspective to think}''.

This leads to modern LLMs performing poorly on subjective tasks challenging to be quantified or measured objectively~\cite{jentzsch-kersting-2023-chatgpt,wachowiak-gromann-2023-gpt,mao2023gpteval}. Moreover, due to the nature of the subjective tasks, identifying a chain-of-thought pathway similar to that in conventional reasoning tasks is difficult~\cite{ric}.  For instance,  the generated reasoning pathways can even mislead the model to provide incorrect answers, as examples shown in Figure~\ref{fig:challenge}. Therefore, directly exploring CoT prompting techniques may not be practical for subjective tasks, motivating us to propose a general method to enhance the ability of LLMs to solve various subjective tasks.

Inspired by the theory of mind ~\cite{Premack_Woodruff_1978,Wellman}, which refers to the ability to attribute mental states to oneself and others and to understand that these mental states can influence behavior,  we leverage different perspectives to better address the aforementioned challenging subjective tasks rather than directly answering the questions based on the LLMs' own direct perspective (e.g., zero-shot ~\cite{gpt3} or zero-shot-CoT reasoning ~\cite{kojima2023large}.  As shown in Figure~\ref{fig:3perspectives}, we categorize reasoning methods of LLMs into three perspectives: 1) direct perspective, which involves the model directly answering questions or tasks based on its internal understanding without considering external factors or alternative viewpoints; 2) role perspective, which focuses on assigning specific roles to the model, simulating different viewpoints or expertise within a given context or scenario; 3) third-person perspective, which involves the model considering external viewpoints or perspectives beyond its own, similar to how a third party or observer might view a situation.

Based on the intuition that LLMs perform optimally when their operational parameters align with confidence levels in specific contexts, we introduce Reasoning
through Perspective Transition (RPT) method to dynamically select suitable perspectives to solve specific problems. To enable dynamic perspective transitions in reasoning, we adopt in-context learning ~\cite{gpt3}, providing templates for multi-perspective answers. The model then evaluates confidence levels ~\cite{confidence1,confidence2,selftrain} and selects the perspective with the highest confidence. This adaptability allows it to handle diverse subjective tasks more effectively than static methods.

We conduct experiments on four LLMs (including two closed-source models GPT-3.5 ~\cite{openaichatgpt}/GPT-4 ~\cite{openaigpt4}, and two open-source models Llama-3 ~\cite{dubey2024llama}/Qwen-2 ~\cite{yang2024qwen2} across 12 subjective tasks. Extensive experimental results demonstrates that, compared to previous methods based on a single perspective or some simple ensemble-based methods, our approach can improve the performance consistently.

\section{Related Work}
\textbf{Subjective Tasks in NLP.} Compared with \textit{objective} tasks such as commonsense reasoning ~\cite{talmor-etal-2019-commonsenseqa} and arithmetic question-answering ~\cite{cobbe2021gsm8k}, research on LLMs in \textit{subjective} tasks (\textit{e.g.}, metaphor recognition and dark humor detection) ~\cite{rottger-etal-2022-two,kanclerz-etal-2023-pals,sun2023aligning} is still underexplored. Different from objective tasks that can often be clearly defined and solved, subjective tasks involve the capability to perceive context, language nuances, and emotions, which cannot be easily quantified or objectively measured, thereby posing challenges for current LLMs ~\cite{jentzsch-kersting-2023-chatgpt,wachowiak-gromann-2023-gpt,mao2023gpteval}. For example, as shown in results of BigBench\cite{srivastava2023beyond}, the zero-shot accuracy of PaLM-535B ~\cite{chowdhery2023palm} model on metaphor recognition, dark humor detection, and sarcasm detection tasks does not exceed 50\%.

\begin{figure*}[t!]
    \centering
    \includegraphics[width=1.9\columnwidth]{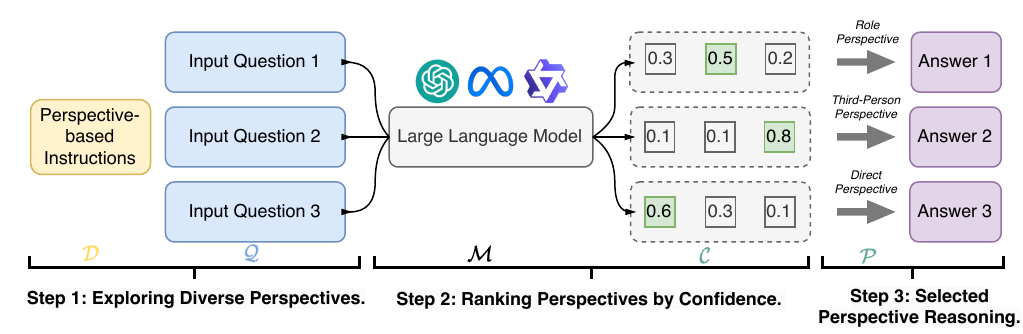}
    \caption{ An overview of RPT pipeline. For each input question, RPT explores the available perspectives and then ranks them based on confidence. Accordingly, the input question is reasoned using the selected perspective.}
    \label{fig:intro}
    \vspace{-1em}
\end{figure*}

\noindent\textbf{In-Context Learning of LLMs.} 
As model parameters increase, LLMs gain stronger in-context learning \cite{gpt3}, enhancing zero-shot and few-shot reasoning without fine-tuning. Chain-of-thought (CoT) prompting \cite{cot, kojima2023large} is widely used to elicit reasoning by adding explicit reasoning steps, emphasizing “\textit{how to think deeper}” \cite{openaio1, deepseek2025r1} while neglecting “\textit{who should think}.” However, recent research \cite{sprague2024cot} suggests that such reasoning pathways are primarily effective for mathematical and symbolic reasoning.
Our work also relies on in-context learning, however, we propose a method based on dynamic perspective transition to elicit knowledge from the different perspective of LLMs, which does not rely on a single reasoning pathway and achieve better results on a wider range of subjective tasks. Our work also relates to LLM-based multi-agents ~\cite{xi2023rise,wang2024survey}, aiming to enhance LLMs' ability to understand context, analyze problems, and generate solutions beyond a single fixed perspective.

\noindent\textbf{Perspective Transition of LLMs.} There are various ways to use LLMs currently that are based on different perspectives: 1) Direct prompting methods ~\cite{gpt3,cot,kojima2023large} let the model to provide answers based on the factual knowledge or reasoning ability by LLMs themselves directly, without setting specific roles. 2) By assigning roles ~\cite{xu2023expertprompting,wang-etal-2024-incharacter,thinktwice} such as experts and engaging in role-playing dialogue, the internal knowledge of LLMs on specific roles can be elicited. 3) By constructing scenarios through multi-agent cooperation ~\cite{wang2023unleashing,deem}, debates ~\cite{du2023mdebates}, or dialogues ~\cite{ric}, and then providing answers from a third-person perspective by incorporating contextualized information by the constructed agents. The previous methods only consider a fixed perspective and validate the effectiveness in certain problems. In contrast, RPT is orthogonal to improvements such as deep long-chain CoT~\cite{deepseek2025r1}, enabling LLMs to adaptively select the most suitable perspective to solve various subjective tasks, which has not been studied in previous research.


\section{Method}

\begin{table}[t!]
    \centering
    \resizebox{\columnwidth}{!}{
    \begin{tabular}{l}
    \toprule
    \textbf{Illustration of the Unified Prompting and Response} \\
    \midrule
    \textcolor{gray}{/* Input */} \\
    \makecell[l{p{8cm}}]{\colorbox{cream}{\textcolor{black}{/* Perspective-based Instructions */}}\\
   Here are some examples of perspectives you can choose:\\   
Direct Perspective (answer the question directly).\\
.......\\
Role Perspective (assume you are some roles (e.g., expert) and answer the question).\\
.......\\
Third-person Perspective (simulate some agents and discussions, then summarize and give the answer).\textcolor{gray}{($\mathcal{D}$)}}\\
.......\\

    \makecell[l{p{8cm}}]{\colorbox{LightSkyBlue}{\textcolor{black}{/* Input Question */}}\\
    Metaphor Recognition: Identify whether the sentence ``Adam did not understand the root of the crisis'' is a paraphrase of the metaphoric sentence ``Adam did not understand the solution to the crisis''~\textcolor{gray}{($\mathcal{Q}$)}\\
    \\
    Firstly, analyzing the question from diverse perspectives, and \textcolor{black}{selecting some available perspectives based on the question.}~\textcolor{gray}{($\mathcal{T}_1$)}\\
    Secondly, \textcolor{black}{ranking all the perspectives according to the confidence level in solving the question}. [Note: The confidence indicates how likely you think your answer is true].~\textcolor{gray}{($\mathcal{T}_2$)}\\
    Finally, choosing the method with the highest confidence to solve the question, \textcolor{black}{reason and give the final answer}. Select from True or False.~\textcolor{gray}{($\mathcal{T}_3$)}\\
    \\
    \textcolor{gray}{/* Response of $\mathcal{M}$ */} \\
    \textcolor{teal}{Perspective and Confidence:}~\textcolor{gray}{($\mathcal{P}$, $\mathcal{C}$)}\\
    \textcolor{teal}{Third-person Perspective, 85\%}\\
    \textcolor{teal}{Role Perspective, 70\%} \\
    \textcolor{teal}{Direct Perspective, 60\%}\\
    \\
    \textcolor{Brown}{Selected Perspective Reasoning:}~\textcolor{gray}{($\mathcal{R}$)}\\
    \textcolor{Brown}{Tom: Adam did not understand the root of the crisis. }\\
    \textcolor{Brown}{Jerry: So, does that mean Adam did not understand the solution to the crisis?}\\
    \textcolor{Brown}{Tom: Not exactly. Understanding the root of the crisis doesn't necessarily mean understanding the solution. Understanding the root is figuring out what causes the problem, while a solution refers to ways of solving the problem. They can be related but are not synonymous.}\\
    \textcolor{Brown}{Answer: False}
    }
    \\
    \bottomrule
    \end{tabular}}
    \caption{The example of a metaphor recognition task, a specific question, the unified prompt, and the corresponding response.}
    \label{table:unifiedprompt}
    \vspace{-1.2em}
\end{table}

The overall pipeline of the proposed RPT is structured into three steps, as shown in Figure~\ref{fig:intro}. Firstly, we input the task description and a specific question, prompting the model to select the most appropriate perspective for answering the question. Secondly, the model evaluates and ranks these perspectives based on their confidence levels in addressing the question. Thirdly, the model adopts the perspective with the highest confidence to formulate and deliver the definitive answer.
  
Formally, given a subjective task with a description $\mathcal{D}$ and a specific question $\mathcal{Q}$, our goal is to let the LLM $\mathcal{M}$ solve the question $\mathcal{Q}$.

\noindent \textbf{Step 1: Exploring Diverse Available Perspectives.} We first let LLM $\mathcal{M}$ explore diverse perspectives $\mathcal{P}$ according to the description $\mathcal{D}$ and question $\mathcal{Q}$, promoting more comprehensive reasoning. Specifically, we define: 
\begin{equation}
    \mathcal{P}=\{p_{1},p_{2},..., p_{n}\}=\mathcal{M}(\mathcal{D}\oplus\mathcal{Q}\oplus\mathcal{T}_1),
\end{equation}
where $n$ (the number of perspectives) is usually $3$, $\oplus$ denotes concatenation operation. $\mathcal{T}_1$ is a prompt serving as a trigger sentence, for example, we can set $\mathcal{T}_1$ as ``\textit{Firstly, analyzing the question from diverse perspectives, and selecting some available perspectives based on the question}''.

\noindent \textbf{Step 2: Ranking Perspectives by Confidence Level.} Then, based on the generated perspectives, we allow the LLM $\mathcal{M}$ to rank them according to their confidence levels $\mathcal{C}$ in solving the question.
\begin{equation}
    \mathcal{P},\mathcal{C}=\mathcal{M}(\mathcal{D}\oplus\mathcal{Q}\oplus\mathcal{T}_2),
\end{equation}
where $\mathcal{T}_2$ is a prompt for ranking the confidence level of all the available perspectives. For example, we can set $\mathcal{T}_2$ as ``\textit{Secondly, Ranking all the methods according to the confidence level in solving the question. [Note: The confidence indicates how likely you think your answer is true.]}''.

\noindent \textbf{Step 3: Selected Perspective Reasoning.} Finally, we take the original task description $\mathcal{D}$, question $\mathcal{Q}$, and the ranked confidence level perspective $\mathcal{P}$ as the input, letting LLM $\mathcal{M}$ give the final response $\mathcal{R}$:
\begin{equation}
    \mathcal{R}=\mathcal{M}(\mathcal{D}\oplus\mathcal{Q}\oplus\mathcal{P}\oplus\mathcal{T}_3),
\end{equation}
where $\mathcal{T}_3$ is the last prompt leading to the final answer which can be set as ``\textit{Finally, Choosing the perspective with the highest confidence to solve the question, and give the final answer}''.

\noindent \textbf{Combine All Steps through Unified Prompting.} In practice, we find that the three aforementioned steps can be combined and accomplished through a single prompt $\mathcal{T}$. In this way, our method only requires inference once through the LLM to obtain the answer to the question:
\begin{equation}
\begin{array}{l}
\begin{aligned}
    \mathcal{T}&=\mathcal{T}_1\oplus\mathcal{T}_2\oplus\mathcal{T}_3,\\
    \mathcal{P},\mathcal{C},\mathcal{R}&=\mathcal{M}(\mathcal{D}\oplus\mathcal{Q}\oplus\mathcal{T}),
\end{aligned}
\end{array}
\label{eq:unify}
\end{equation}
where an example of the unified prompt and response is shown in Table~\ref{table:unifiedprompt}.

\section{Experiments}

\subsection{Settings}

\noindent\textbf{Datasets}. We evaluate the effectiveness of our method on twelve subjective reasoning datasets, which can be categorized into five types, as shown in Table~\ref{tab:number_dataset}. Notably, for SemEval and cultural-related datasets which contain training sets, we evaluate in both zero-shot and few-shot settings. For the other tasks, we utilize corresponding test sets from BigBench\footnote{\url{https://github.com/google/BIG-bench/tree/main/bigbench/benchmark_tasks/}} ~\cite{srivastava2022beyond} and only evaluate in zero-shot settings.

\begin{table}[t!]
\centering
    \resizebox{1.0\linewidth}{!}{
    \begin{tabular}{lcr}
    \toprule 
    \textbf{Dataset} (names in short)&\textbf{Subjective Tasks}&\textbf{\#Train/Dev/Test}\\
         \midrule
         (\textit{Linguistic Rhetoric})&&\\
         \textbf{Metaphor} ~\cite{mohler-etal-2016-introducing}&Metaphor Understanding&-/-/$680$\\
         \textbf{SNARKS} ~\cite{khodak-etal-2018-large}&Sarcasm Detection&-/-/$181$\\
         \textbf{Humor} ~\cite{hoffmann2022training}&Dark Humor Detection&-/-/$80$\\
         \midrule
         (\textit{Disambiguation QA})&&\\
         \textbf{Pronoun} ~\cite{rudinger-etal-2018-gender}&Pronoun Resolution&-/-/$258$\\
         \textbf{Anachronisms} ~\cite{geva-etal-2021-aristotle}&Identifying Anachronisms&-/-/$230$\\
         \midrule
         (\textit{Stance Detection})&&\\
         \textbf{SEQ} ~\cite{hendrycks2020aligning}&Simple Ethical Questions&-/-/$115$\\
        \textbf{SemEval} ~\cite{semeval2016} &Opinion Analysis&$2$,$194$/$621$/$707$\\
        \midrule
        (\textit{Cultural-Related})&&\\
        \textbf{SocNorm} ~\cite{ch-wang-etal-2023-sociocultural}&Sociocultural Norm NLI&$2$,$301$/$300$/$768$\\
         \textbf{e-SocNorm} ~\cite{ch-wang-etal-2023-sociocultural}&Sociocultural Norm NLI&$2$,$301$/$300$/$768$\\
         \textbf{CALI} ~\cite{huang-yang-2023-culturally}&Culturally Aware NLI&$1$,$757$/$-$/$440$\\
         \midrule
         (\textit{Traditional NLI})&&\\
         \textbf{Entailment} ~\cite{srivastava2022beyond}&Analytic Entailment&-/-/$70$\\
         \textbf{IPA} ~\cite{williams-etal-2018-broad}&NLI in the International Phonetic Alphabet&-/-/$126$\\
        \bottomrule
    \end{tabular}}
    \caption{Statistics and resources of datasets.}
    \label{tab:number_dataset}
    \vspace{-1em}
\end{table}

\noindent\textbf{Baselines.}
We compare our method with 11 baselines including different single perspective methods and ensemble-based methods as follows.

\begin{table*}[t!]
	\centering
    \scalebox{0.55}{
	\begin{tabular}{clccccccccccccc}
 \toprule
        \multirow{3.5}*{\bf{Type}}&\multirow{3.5}*{\bf{Method}}&\multicolumn{3}{c}{\bf{Linguistic Rhetoric}}&\multicolumn{2}{c}{\bf{Disambiguation QA}}&\multicolumn{2}{c}{\bf{Stance Detection}}&\multicolumn{3}{c}{\bf{Cultural-Related}}&\multicolumn{2}{c}{\bf{Traditional NLI}} &\multirow{3.5}*{\bf{\textsc{Avg.}}}\\
        \cmidrule(lr){3-5}\cmidrule(lr){6-7}\cmidrule(lr){8-9}\cmidrule(lr){10-12}\cmidrule(lr){13-14}
        &&\textbf{Metaphor}&\textbf{SNARKS}&\textbf{Humor}&\textbf{Pronoun}&\textbf{Anach.}&\textbf{SEQ}&\textbf{SemEval}&\textbf{SocNorm}&\textbf{e-SocNorm}&\textbf{CALI}&\textbf{Entail.}&\textbf{IPA}\\
        &&(Acc.)&(Acc.)&(Acc.)&(Acc.)&(Acc.)&(Acc.)&(F1)&(F1)&(F1)&(Acc.)&(Acc.)&(Acc.)&\\
    \midrule
     -&\textit{Random}&$50.00$&$50.00$&$50.00$&$33.33$&$50.00$&$25.00$&$50.00$&$33.33$&$33.33$&$33.33$&$50.00$&$33.33$&$40.97$\\
    -&\textit{Majority}&$61.62$&$53.59$&$50.00$&$30.23$&$50.00$&$10.43$&$\phantom{0}0.00$&$\phantom{0}0.00$&$\phantom{0}0.00$&$38.09$&$57.14$&$38.89$&$32.50$\\
    \midrule
    
    \multicolumn{15}{c}{ (\texttt{Llama-3-8b-instruct})} \\
S1&Direct Prompt ~\cite{gpt3}&$66.03$&$58.56$&$60.00$&$43.41$&$50.00$&$61.74$&$71.00$&$39.15$&$48.49$&$42.95$&$51.43$&$39.68$&$52.70$\\ 
S1&Zero-Shot-CoT ~\cite{kojima2023large}&$67.06$&$\bf{70.72}$&$63.75$&$46.90$&$61.74$&$73.04$&$72.45$&$40.07$&$52.84$&$47.95$&$54.29$&$44.44$&$57.94$\\
S2&Role-Play Prompting ~\cite{kong-etal-2024-better}&$65.00$&$64.09$&$\underline{65.00}$&$45.35$&$53.91$&$72.17$&$73.26$&$51.36$&$56.44$&$46.82$&$51.54$&$43.65$&$57.38$\\
S3&Reason in Conversation ~\cite{ric}&$\underline{76.32}$&$\underline{69.72}$&$58.75$&$48.06$&$52.17$&$80.00$&$74.71$&$48.15$&$\bf{64.05}$&$48.86$&$\bf{58.57}$&$50.79$&$\underline{60.85}$\\
E&Ensemble ~\cite{ensemble2}&$68.09$&$64.64$&$50.00$&$37.60$&$\underline{69.57}$&$\underline{82.61}$&$\underline{77.00}$&$44.60$&$58.72$&$\bf{54.77}$&$57.14$&$53.97$&$59.89$\\
E&Reranking ~\cite{reranking1}&$71.47$&$58.56$&$52.50$&$\underline{51.78}$&$59.13$&$72.17$&$72.23$&$52.86$&$52.34$&$48.41$&$55.71$&$\underline{54.76}$&$59.49$\\
E&CoT-SC ~\cite{sccot}&$65.88$&$48.07$&$\bf{66.25}$&$45.74$&$61.30$&$78.26$&$76.01$&$\bf{55.86}$&$59.64$&$47.73$&$52.86$&$\bf{61.90}$&$59.96$\\
D&\textbf{RPT (Ours)}&$\bf{81.76}$&$60.22$&$\underline{65.00}$&$\bf{53.49}$&$\bf{72.17}$&$\bf{89.57}$&$\bf{77.44}$&$\underline{53.52}$&$\underline{61.72}$&$\underline{51.59}$&$\bf{58.57}$&$44.44$&$\bf{64.12}$\\
    \addlinespace[1pt]
    \cdashline{1-15}
    \addlinespace[3pt]
    \multicolumn{15}{c}{ (\texttt{qwen-2-7b-instruct})} \\
S1&Direct Prompt ~\cite{gpt3}&$79.85$&$61.88$&$60.00$&$56.98$&$64.38$&$86.09$&$70.17$&$38.53$&$47.93$&$42.27$&$58.57$&$58.73$&$60.45$\\ 
S1&Zero-Shot-CoT ~\cite{kojima2023large}&$83.09$&$64.03$&$\underline{63.75}$&$54.65$&$63.48$&$79.13$&$73.36$&$43.99$&$47.79$&$46.59$&$62.86$&$62.70$&$62.12$\\
S2&Role-Play Prompting ~\cite{kong-etal-2024-better}&$78.97$&$65.75$&$56.25$&$52.25$&$60.87$&$86.96$&$72.25$&$46.77$&$51.21$&$49.77$&$64.29$&$57.14$&$61.87$\\
S3&Reason in Conversation ~\cite{ric}&$80.59$&$69.61$&$60.00$&$60.47$&$63.91$&$87.83$&$75.06$&$\underline{49.57}$&$56.40$&$53.18$&$64.29$&$60.32$&$65.10$\\
E&Ensemble ~\cite{ensemble2}&$\underline{86.03}$&$\underline{74.59}$&$62.50$&$54.65$&$63.48$&$88.70$&$72.66$&$46.08$&$58.02$&$53.64$&$\underline{71.43}$&$\bf{66.67}$&$\underline{66.54}$\\
E&Reranking ~\cite{reranking1}&$83.23$&$72.38$&$60.00$&$54.65$&$62.17$&$87.04$&$74.21$&$44.81$&$54.18$&$\bf{54.32}$&$\bf{74.29}$&$\underline{65.08}$&$65.53$\\
E&CoT-SC ~\cite{sccot}&$\bf{86.32}$&$\bf{79.56}$&$47.50$&$\bf{66.28}$&$\bf{70.00}$&$\underline{92.17}$&$\underline{75.29}$&$44.44$&$\underline{61.90}$&$\bf{54.32}$&$50.00$&$56.35$&$65.34$\\
D&\textbf{RPT (Ours)}&$84.41$&$69.61$&$\bf{65.00}$&$\underline{63.95}$&$\underline{68.70}$&$\bf{94.78}$&$\bf{76.58}$&$\bf{51.02}$&$\bf{68.29}$&$52.27$&$67.14$&$61.90$&$\bf{68.64}$\\
    \addlinespace[1pt]
    \cdashline{1-15}
    \addlinespace[3pt]
     \multicolumn{15}{c}{ (\texttt{gpt-3.5-turbo-1106})} \\
 S1&Direct Prompt ~\cite{gpt3}&$85.74$&$77.35$&$58.75$&$55.04$&$70.43$&$75.65$&$71.30$&$43.25$&$45.27$&$52.94$&$60.00$&$50.79$&$62.21$\\
     S1&Zero-Shot-CoT ~\cite{kojima2023large}&$86.47$&$78.45$&$57.50$&$60.47$&$64.78$&$72.17$&$73.79$&$44.68$&$51.53$&$52.75$&$58.57$&$55.56$&$63.06$\\
     S2&Role-Play Prompting ~\cite{kong-etal-2024-better}&$82.64$&$77.40$&$57.25$&$60.39$&$71.74$&$78.39$&$71.10$&$47.61$&$49.13$&$55.68$&$61.43$&$57.14$&$64.16$\\
     S3&Reason in Conversation ~\cite{ric}&$\underline{87.94}$&$82.32$&$\bf{71.25}$&$\underline{62.79}$&$72.61$&$81.74$&$74.27$&$56.02$&$59.98$&$57.27$&$62.86$&$57.14$&$68.85$\\  
     E&Ensemble ~\cite{ensemble2}&$84.26$&$76.80$&$66.25$&$59.61$&$72.17$&$86.26$&$70.32$&$48.25$&$56.51$&$52.95$&$64.29$&$65.08$&$66.90$\\ 
     E&Reranking ~\cite{reranking1}&$81.76$&$79.56$&$65.00$&$54.65$&$72.17$&$81.30$&$77.27$&$51.18$&$63.99$&$\underline{60.91}$&$61.42$&$63.49$&$67.73$\\ 
    E&CoT-SC ~\cite{sccot}&$84.85 $ & $\underline{86.74} $ & $67.50 $ & $47.29 $ & $\bf{74.35} $ & $\underline{92.17} $ & $\underline{81.18} $ & $\underline{59.35} $ & $\underline{65.53} $ & $59.09 $ & $\underline{87.14} $ & $\underline{75.40} $ & $\underline{73.38}$\\
D&\textbf{RPT (Ours)}&$\bf{91.76} $ & $\bf{87.29} $ & $\underline{70.00} $ & $\bf{65.12} $ & $\underline{73.48} $ & $\bf{99.13} $ & $\bf{81.43} $ & $\bf{59.81} $ & $\bf{77.57} $ & $\bf{61.13} $ & $\bf{88.57} $ & $\bf{80.00} $ & $\bf{77.94}$\\
 \addlinespace[1pt]
    \cdashline{1-15}
    \addlinespace[3pt]
 \multicolumn{15}{c}{ (\texttt{gpt-4-0613})} \\
S1&Direct Prompt ~\cite{gpt3}&$94.85$&$86.19$&$65.00$&$72.09$&$82.17$&$92.17$&$72.78$&$45.31$&$46.81$&$60.40$&$68.57$&$75.40$&$71.81$\\
     S1&Zero-Shot-CoT ~\cite{kojima2023large}&$\underline{95.88}$&$87.29$&$66.25$&$69.38$&$80.00$&$93.91$&$75.47$&$48.74$&$47.45$&$60.90$&$75.71$&$73.02$&$72.83$\\
     S2&Role-Play Prompting ~\cite{kong-etal-2024-better}&$93.97$&$82.87$&$63.75$&$67.05$&$80.87$&$96.52$&$73.71$&$52.31$&$54.51$&$58.86$&$77.14$&$73.81$&$72.95$\\
S3&Reason in Conversation ~\cite{ric}&$95.29$&$\underline{92.27}$&$\bf{67.50}$&$\underline{75.58}$&$\underline{86.96}$&$95.65$&$\underline{76.34}$&$\underline{58.27}$&$61.12$&$61.13$&$87.14$&$80.95$&$\underline{78.18}$\\
E&Ensemble ~\cite{ensemble2}&$95.44$&$88.95$&$65.00$&$61.63$&$81.74$&$\underline{98.26}$&$75.57$&$58.33$&$\underline{66.78}$&$\underline{63.18}$&$87.14$&$78.57$&$76.72$\\
E&Reranking ~\cite{reranking1}&$94.71$&$84.53$&$65.00$&$65.89$&$81.73$&$97.39$&$74.70$&$56.28$&$66.18$&$59.32$&$\bf{88.57}$&$76.19$&$75.87$\\
E&CoT-SC ~\cite{sccot}&$\bf{96.00} $ & $84.53 $ & $73.75 $ & $73.26 $ & $83.04 $ & $\bf{99.13} $ & $72.52 $ & $53.26 $ & $66.23 $ & ${62.95} $ & $57.14 $ & $\underline{83.33} $ & $75.43$\\
D&\textbf{RPT (Ours)}&$95.29 $ & $\bf{92.82} $ & $\bf{67.50} $ & $\bf{75.97} $ & $\bf{87.39} $ & $97.39 $ & $\bf{78.53} $ & $\bf{61.78} $ & $\bf{75.87} $ & $\bf{63.64} $ & $\bf{88.57} $ & $\bf{84.92} $ & $\bf{80.81}$\\
     \bottomrule
    \end{tabular} }
  \caption{Zero-shot results. To demonstrate generalizability, we repeat each set of experiments separated by dashed lines on four LLMs independently. \textit{Random} represents the result of random prediction with uniform probability, and \textit{Majority} represents the result of predicting the label with the highest proportion. S1: single direct perspective, S2: single role perspective, S3: single third-person perspective, E: ensemble-based method. D: dynamic perspective. For each dataset, the best result is \textbf{in bold} and the second-best result is \underline{underlined}.}
	\label{table:mainresults}
\end{table*}

 \textit{Single Direct Perspective.} \textbf{Directly Prompt} ~\cite{gpt3} directly use the question as input in zero-shot or few-shot manners. \textbf{ICL} ~\cite{gpt3} (in-context learning) uses examples and labels as few input demonstrations. \textbf{Few-shot-CoT} ~\cite{cot} uses manually created external reasoning pathways as demonstrations. \textbf{Zero-Shot-CoT} ~\cite{kojima2023large} does not rely on demonstrations and elicits the reasoning ability by using ``Let's think step by step.'' as external input. \textbf{Self-Ask} ~\cite{press-etal-2023-measuring} actively proposes and solves subquestions before generating the final answer.

\textit{Single Role Perspective.} \textbf{ExpertPrompt} ~\cite{xu2023expertprompting} introduces the expert identities and customizes information descriptions for LLMs before generating responses. \textbf{Role-Play Prompting} ~\cite{kong-etal-2024-better} also lets models simulate complex human-like interactions and behaviors for zero-shot reasoning.

\textit{Single Third-Person Perspective.} \textbf{SPP} ~\cite{wang2023unleashing} (solo performance prompting) proposes solo performance prompting by involving multi-turn collaboration with multi-persona. \textbf{RiC} ~\cite{ric} (reason in conversation) first lets model generating dialogues between simulated roles, and then summarize conversations and give final answers according to the additional information from conversations.

\textit{Ensemble-based Methods.} \textbf{Ensemble} ~\cite{ensemble1,ensemble2} involves combining multiple model generation to enhance prediction accuracy and robustness (refer to Appendix~\ref{sec:compare_ensemble}).  \textbf{Reranking} ~\cite{reranking1,reranking2} reorder different generation options based on requirements and select the optimal result. \textbf{CoT-SC} ~\cite{sccot} improves performance by generating diverse chain-of-thought reasoning paths and selecting the most self-consistent answer, thereby enhancing the robustness and reliability.

\noindent\textbf{Models.} We evaluate our method on both closed-source models including GPT-4 ~\cite{openaigpt4} and GPT-3.5 ~\cite{openaichatgpt}, and open-source Llama-3 ~\cite{dubey2024llama} and Qwen-2 ~\cite{yang2024qwen2} models. In particular, we use the released API versions of \texttt{gpt-4-0613} and \texttt{gpt-3.5-turbo-1106} by OpenAI, and open-source \texttt{Llama-3-8b-instruct} and \texttt{qwen-2-7b-instruct} models released in Huggingface hub. We set the decoding temperature to $0$ to ensure deterministic outputs and maintain the reproducibility of the responses generated by LLMs.


\begin{table}[t]
\small
    \centering
    \resizebox{1.0\linewidth}{!}{
    \begin{tabular}{clccccc}
    \toprule
          \textbf{Type}&\bf{Method}& \bf{SemEval}&\bf{SocNorm} & \bf{e-SocNorm} & \bf{CALI} &\bf{\textsc{Avg.}}\\
    \midrule
    \multicolumn{7}{c}{ (\texttt{Llama-3-8b-instruct})} \\
    S1&ICL ~\cite{gpt3}&$70.71$&$ 47.82 $&$ 57.73 $&$47.27$&$55.88$  \\
    S1&Few-Shot-CoT ~\cite{gpt3}&$76.45$&$ 48.37 $&$ 57.77 $&$48.41$&$57.75$  \\
    S1&Self-Ask ~\cite{press-etal-2023-measuring}&$76.46$&$ 49.52 $&$ 53.34 $&$48.64$&$56.99$  \\
    S2&ExpertPrompt ~\cite{xu2023expertprompting} &$75.08$&$ 47.46 $&$ 64.85 $&$45.00$&$58.10$  \\
    S3&SPP ~\cite{wang2023unleashing}&$74.91$&$ 40.55 $&$ 56.15 $&$50.68$&$55.57$  \\
    S3&RiC ~\cite{ric}&$77.48$&$ \underline{52.54} $&$ 66.60 $&$50.23$&$61.71$  \\

    E&Ensemble ~\cite{ensemble2}&$76.23$&$ 45.53 $&$ 67.31$&$51.14$&$60.05$  \\
    E&Reranking ~\cite{reranking1}&$71.79$&$ 42.80 $&$ 64.89$&$50.68$&$57.54$  \\
    E&CoT-SC ~\cite{sccot}&$\underline{79.33}$&$ 40.92 $&$ \bf{75.80}$&$\bf{51.59}$&$\underline{61.91}$  \\
    D&\textbf{RPT (Ours)}&$\bf{80.02}$&$ \bf{54.21} $&$ \underline{70.05} $&$\bf{51.59}$&$\bf{63.97}$  \\
    \addlinespace[1pt]
    \cdashline{1-7}
    \addlinespace[3pt]
    \multicolumn{7}{c}{ (\texttt{qwen-2-7b-instruct})} \\
    S1&ICL ~\cite{gpt3}&$70.83$&$ 35.97 $&$ 54.52 $&$52.27$&$53.40$  \\
    S1&Few-Shot-CoT ~\cite{gpt3}&$71.16$&$ 52.01 $&$ 63.51 $&$53.41$&$60.02$  \\
    S1&Self-Ask ~\cite{press-etal-2023-measuring}&$74.09$&$ 47.89 $&$ 56.28 $&$52.05$&$57.58$  \\
    S2&ExpertPrompt ~\cite{xu2023expertprompting} &$72.65$&$ 54.56 $&$ 62.70 $&$52.27$&$60.55$  \\
    S3&SPP ~\cite{wang2023unleashing}&$72.76$&$ 47.89 $&$ 57.91 $&$54.59$&$58.29$  \\
    S3&RiC ~\cite{ric}&$\bf{76.37}$&$ \underline{55.69} $&$ \underline{68.12} $&$55.91$&$\underline{64.02}$  \\
    E&Ensemble ~\cite{ensemble2}&$\underline{75.94}$&$ 29.18 $&$ 44.25$&$\underline{57.73}$&$51.78$  \\
    E&Reranking ~\cite{reranking1}&$71.96$&$ 51.32 $&$ 67.37$&$52.73$&$60.85$  \\
    E&CoT-SC ~\cite{sccot}&$72.55$&$ 30.93 $&$ 54.25$&$\bf{58.64}$&$54.09$  \\
    D&\textbf{RPT (Ours)}&$74.23$&$ \bf{59.73} $&$ \bf{72.52} $&$56.82$&$\bf{65.83}$  \\
    \addlinespace[1pt]
    \cdashline{1-7}
    \addlinespace[3pt]
   \multicolumn{7}{c}{ (\texttt{gpt-3.5-turbo-1106})} \\
    S1&ICL ~\cite{gpt3}&$72.02$&$ 52.95 $&$ 55.60 $&$54.77$&$58.84$  \\
    S1&Few-Shot-CoT ~\cite{gpt3}& $72.06$&$ 53.44 $&$ 61.35 $&$54.55$ &$60.35$\\
    S1&Self-Ask ~\cite{press-etal-2023-measuring}& $73.04$&$53.94 $&$57.81 $&$57.27$ &$60.52$ \\
    S2&ExpertPrompt ~\cite{xu2023expertprompting} &$75.22$ &$46.08 $&$ 65.29$&$55.45$ &$60.51$ \\
    S3&SPP ~\cite{wang2023unleashing}&$72.74$&$ 51.92$&$ 62.01$&$55.91$ &$60.65$ \\
    S3&RiC ~\cite{ric}&$\underline{78.21}$ &$57.70$& $72.78$ & $\bf{60.00}$& $\underline{67.17}$   \\
    E&Ensemble ~\cite{ensemble2}&$77.33$&$ 57.42 $&$ 66.52 $&$ 58.72 $&$65.00$  \\
    E&Reranking ~\cite{reranking1}&$68.73$&$ 50.90 $&$ \underline{74.45}$&$54.32$&$62.10$  \\
    E&CoT-SC ~\cite{sccot}&$74.56$&$ \underline{58.25} $&$ 73.83$&$59.09$&$66.43$  \\
    D&\textbf{RPT (Ours)}&$\bf{80.78}$ &$\bf{62.70}$& $\bf{74.60}$ & $\bf{60.00}$& $\bf{69.52}$   \\
    \addlinespace[1pt]
    \cdashline{1-7}
    \addlinespace[3pt]
    \multicolumn{7}{c}{ (\texttt{gpt-4-0613})} \\
    S1&ICL ~\cite{gpt3}&$73.72$ &$ 54.71 $&$61.41 $&$ 62.50$&$63.09$\\
    S1&Few-Shot-CoT ~\cite{gpt3}&$76.59$&$64.08 $&$67.88 $&$64.77$&$68.33$ \\
    S1&Self-Ask ~\cite{press-etal-2023-measuring}&$73.52$&$56.74 $&$ 64.62$&$65.45$&$65.08$ \\
    S2&ExpertPrompt ~\cite{xu2023expertprompting}&$77.65$&$56.84 $&$68.72 $&$59.77$ &$65.75$\\
    S3&SPP ~\cite{wang2023unleashing}&$78.72$ &$57.74 $&$65.04 $&$54.32$ &$63.96$\\
    S3&RiC ~\cite{ric}&$\underline{80.01}$ &$\underline{66.59}$&${74.45}$&$\underline{65.68}$&$\underline{71.68}$ \\
    E&Ensemble ~\cite{ensemble2}&$68.95$&$ 63.95 $&$ 67.88$&$64.77$&$66.39$  \\
    E&Reranking ~\cite{reranking1}&$66.61$&$ 60.52 $&$ 72.45$&$62.05$&$65.41$  \\
    E&CoT-SC ~\cite{sccot}&$69.85$&$ 63.37 $&$ \underline{75.37}$&$57.05$&$66.41$  \\
    D&\textbf{RPT (Ours)}&$\bf{80.11}$ &$\bf{66.79}$&$\bf{79.89}$&$\bf{66.59}$&$\bf{73.35}$ \\
    \bottomrule
    \end{tabular}
    }

\caption{Main results of baselines and our RPT method in few-shot settings. S1: single direct perspective, S2: single role perspective, S3: single third-person perspective, E: ensemble-based method. D: dynamic perspective. We select the same $3$-shot demonstrations from the training sets to each method for fair comparison.}    
\label{table:mainresults2}
\vspace{-0.5em}
\end{table}


\subsection{Zero-shot Results}

In Table~\ref{table:mainresults}, we shows the experimental results of the baselines and our RPT method in zero-shot settings. From the experimental results, we can observe that:

\noindent \textbf{RPT method consistently outperforms the baselines in most settings.} Due to its ability to rank perspectives and select different perspectives to suit various subjective scenarios, our method achieves an average improvement of 3.27 points on all subjective tasks using the open-source model Llama-3 compared to the best-performing baseline. Similarly, on the closed-source GPT-3.5 model, our method achieves an average improvement of 4.56 points. Since subjective tasks vary widely, RPT achieves optimal performance through dynamic selection. For instance, on the Metaphor dataset, which requires complex contextual subjective understanding, our method, using Llama-3, outperforms the RiC method, which focuses on dialogue understanding, by 5.44 points.

\noindent \textbf{Compared to baseline methods, our RPT method exhibits greater robustness.} Although baselines introduce different perspectives to adapt to subjective tasks, they are typically effective only in specific domains. For example, using Llama-3, the Zero-Shot-CoT baseline achieves good performance on the Linguistic Rhetoric task, reaching the highest 70.72 accuracy on the SNARKS dataset, but performs poorly on tasks requiring complex contexts and cultural backgrounds, such as stance detection and culturally related datasets. For example, it only achieves 40.07 F1 score on SocNorm, the lowest among all baselines. Conversely, the RiC baseline, which employs role-playing for dialogue simulation, performs well in culturally relevant scenarios, achieving the highest F1 score of 64.05 on the e-SocNorm dataset, but struggles in the Linguistic Rhetoric task. Overall, different baselines that simulate distinct perspectives excel in specific domains but exhibit poor generalizability. In contrast, our method demonstrates consistent improvements across various subjective tasks, making it more robust.

\noindent \textbf{Introducing more diverse perspectives into LLM and switching dynamically among them improves subjective reasoning performance.} The RPT method achieves ensemble through exploring diverse perspectives and ranking perspectives by confidence level. In various settings, baselines that utilize multiple perspectives (e.g., RiC) outperform those that employ a single perspective (e.g., CoT), with scores of 60.85 vs. 58.77 on Llama-3 and 65.10 vs. 62.12 on Qwen. RPT takes this further by proposing dynamic perspective shifts, which offer high generalization and scalability, resulting in optimal performance through dynamic ensemble of all the baselines mentioned above.

\noindent \textbf{Subjective reasoning is highly challenging and performs better on powerful LLMs capable of deep reasoning.}  RPT almost always achieves the best performance on powerful models capable of deep reasoning, such as GPT-3.5 and GPT-4, while being more unstable on weaker open-source models. This suggests that rather than merely enhancing the ability to think from its own perspective, considering different perspectives can further improve LLM performance, especially for well-trained LLMs capable of long-chain deep reasoning. The performance gap between close-source models and open-source models narrows when using the RPT method, given that LLMs still possess the knowledge required for subjective reasoning and not been effectively elicited during training. RPT selects the method best suited to the model's strengths, effectively eliciting LLM capabilities in subjective tasks, and compensating for the lack of subjective task data during LLM training.

\subsection{Few-shot Results}

In Table~\ref{table:mainresults2}, we present the 3-shot results (refer to Section~\ref{sec:number_of_shots} for other settings). Similar to the zero-shot results, RPT method achieves the best average performance across different models. For instance, on Llama-3, RPT surpasses CoT-SC, the best-performing baseline, by 2.06 points.

A possible explanation is that providing a few examples in the prompt generally benefits LLM performance by providing context. However, subjective tasks are not well-defined and directly solvable, leading to significant differences between examples. As a result, LLMs exhibit varying confidence across examples, limiting the performance gains from examples and sometimes introducing noise or bias. For instance, using 3-shot examples in the RiC baseline lead to an average performance drop of 6.50 points on GPT-4. In contrast, RPT choose among perspectives and evaluates confidence for each input and method, providing finer-grained supervision signals and resulting in an average performance gain of 1.67 points.

\begin{figure}[t]
    \centering
    \includegraphics[width=0.49\textwidth]{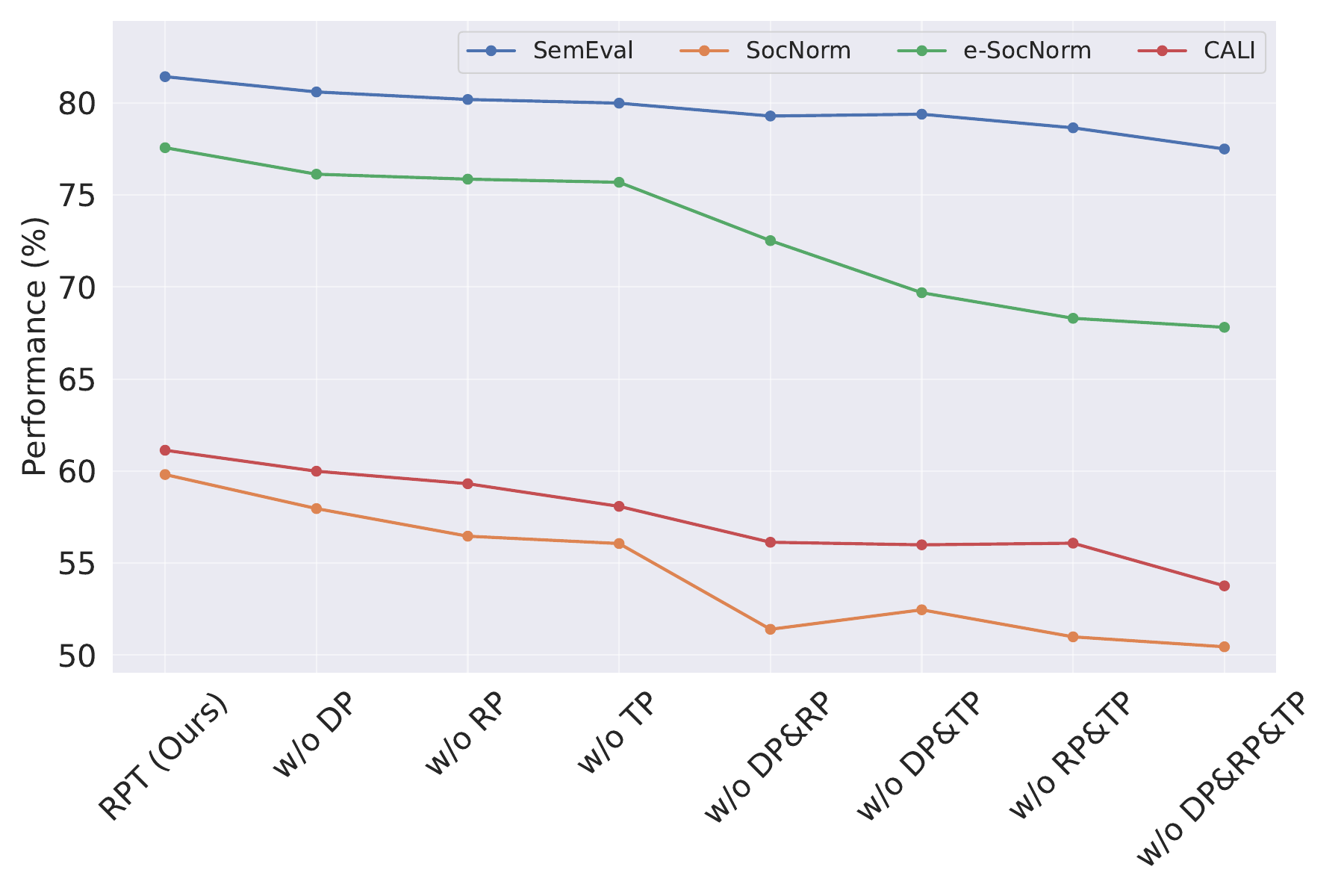}
    \caption{The impact of different perspectives on the RPT method. DP: direct perspective, RP: role perspective, TP: third-person perspective.}
    \label{fig:ablation}
\end{figure}

\begin{figure*}[t!]
    \centering
    \begin{minipage}[t]{0.99\textwidth}
        \centering
        \includegraphics[width=0.99\textwidth]{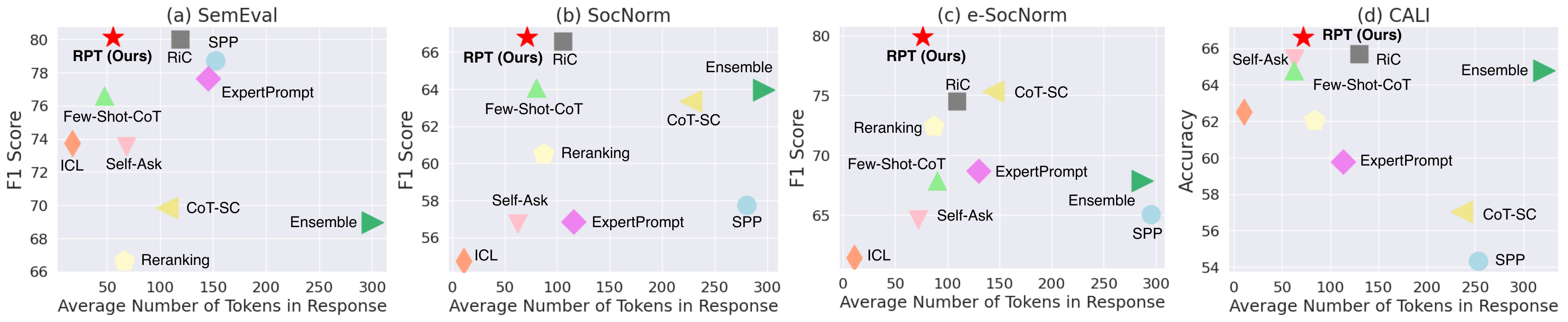}
        \caption{The relationship between the performance and prediction lengths of the 3-shot experiments on GPT-4.}
        \label{fig:length_cor}
    \end{minipage}

    \vspace{1em}
    \begin{minipage}[t]{0.48\textwidth}
        \centering
        \includegraphics[width=0.99\textwidth]{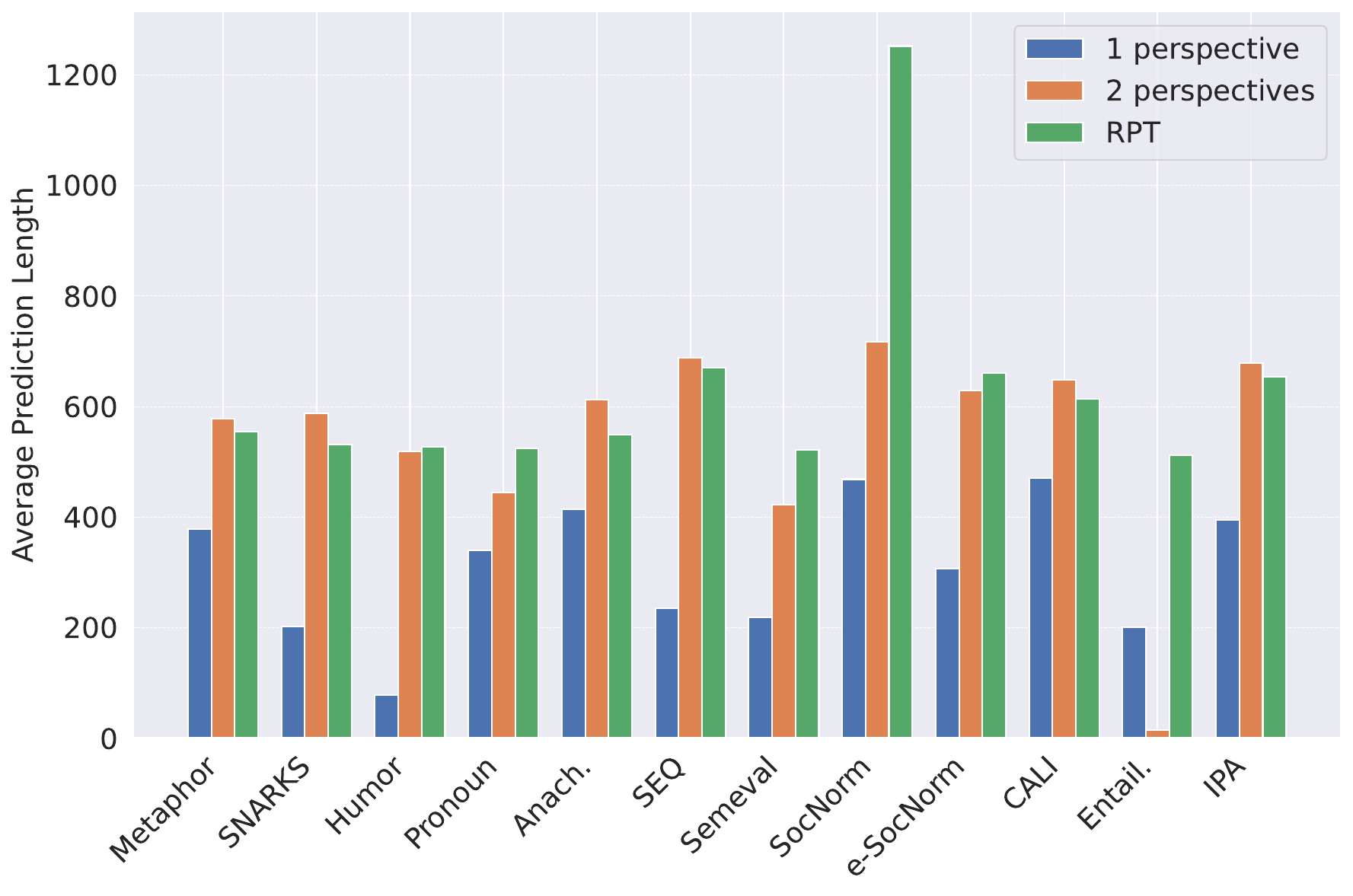}
        \caption{The inference cost of RPT when using different numbers of perspectives. RPT does not significantly increase inference costs on most datasets.}
        \label{tab:length_stat}
    \end{minipage}
    \hspace{0.8em}
    \begin{minipage}[t]{0.48\textwidth}
        \includegraphics[width=0.99\columnwidth]{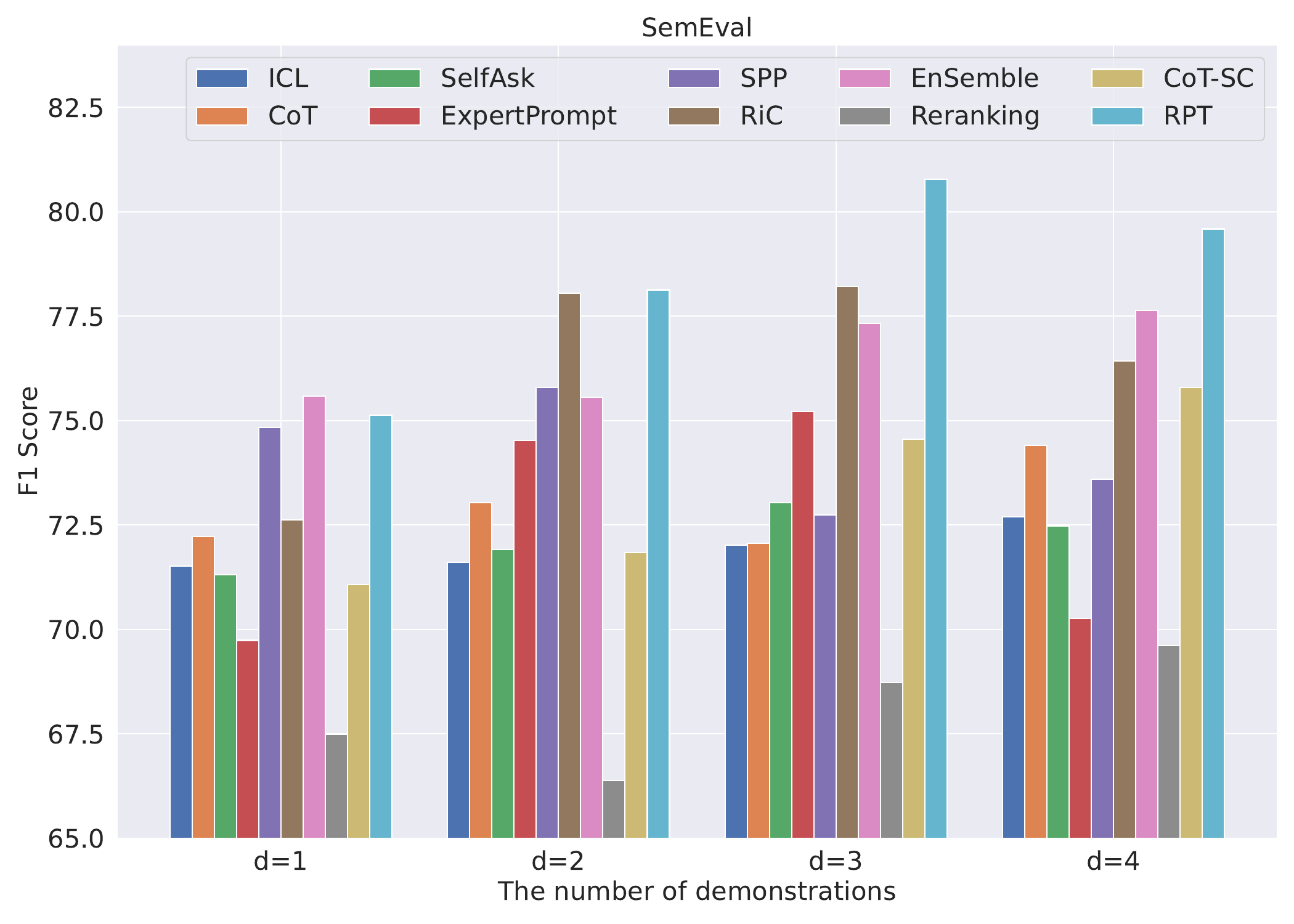}
    \caption{The performance of baselines and our RPT method by using different numbers of demonstrations ($d=1,2,3,4$) in few-shot settings. } 
    \label{fig:fig4}        
    \end{minipage}

\end{figure*}

\section{Analyses and Discussion}

\subsection{Ablation Study}

As shown in Figure~\ref{fig:ablation}, we further investigate the impact of every perspective on the RPT method. The full RPT method achieves the best performance across all datasets. From the results of the ablation study, we can  observe the following:

Removing any single perspective results in an average performance drop of 1.32–2.53 points, indicating that direct perspective, role perspective, and third-person perspective each have unique and irreplaceable contributions to subjective reasoning tasks. Specifically, removing the third-person component has the greatest impact, followed by role perspective and direct perspective, suggesting that the flexibility of switching between perspectives benefits overall performance.

Removing any two perspectives results in an even greater average performance drop of 5.15-6.48 points, and removing all perspectives (i.e., performing simple reasoning) leads to the highest performance drop of 7.60 points. As the number of perspectives removed increases, the range of dynamic switching decreases, causing a corresponding decline in RPT performance. This highlights the crucial role of switching between different perspectives in the RPT method.

In summary, all perspectives involved in RPT and the ability to flexibly switch between them are essential for achieving optimal performance. Thus, every component of our method is effective. The detailed ablation experiment results are presented in Appendix~\ref{app:detail}.

\subsection{Analysis on Performance-Inference Cost}
\label{sec:cost}

Unlike methods such as ensemble and oversampling, which multiply inference costs, RPT aims to improve performance while reducing inference costs.

To estimate the inference cost, we follow \citet{ric} by representing the inference cost by the length of the response during the reasoning process before producing the final answer. In Figure~\ref{fig:length_cor}, using the 3-shot GPT-4 as an example as detailed in Appendix~\ref{sec:shot_number_detail}, we plot the length-performance relationship for RPT and the baselines. It can be observed that compared to most baselines, RPT achieves the best performance with a smaller inference cost, demonstrating the efficiency of the dynamic perspective selection approach. As shown in Figure~\ref{tab:length_stat}, we compare the inference cost of the RPT method with its variants that only allow the use of one or two perspectives. The results indicate that the RPT method incurs a cost similar to that of the variants allowing only one or two perspectives in most datasets, without increasing the inference length, and thus the additional cost is minimal. This is because RPT performs method selection and perspective switching, avoiding the substantial inference cost caused by ensembling multiple perspectives.

\begin{figure*}[t!]
    \centering
    \includegraphics[width=0.96\textwidth]{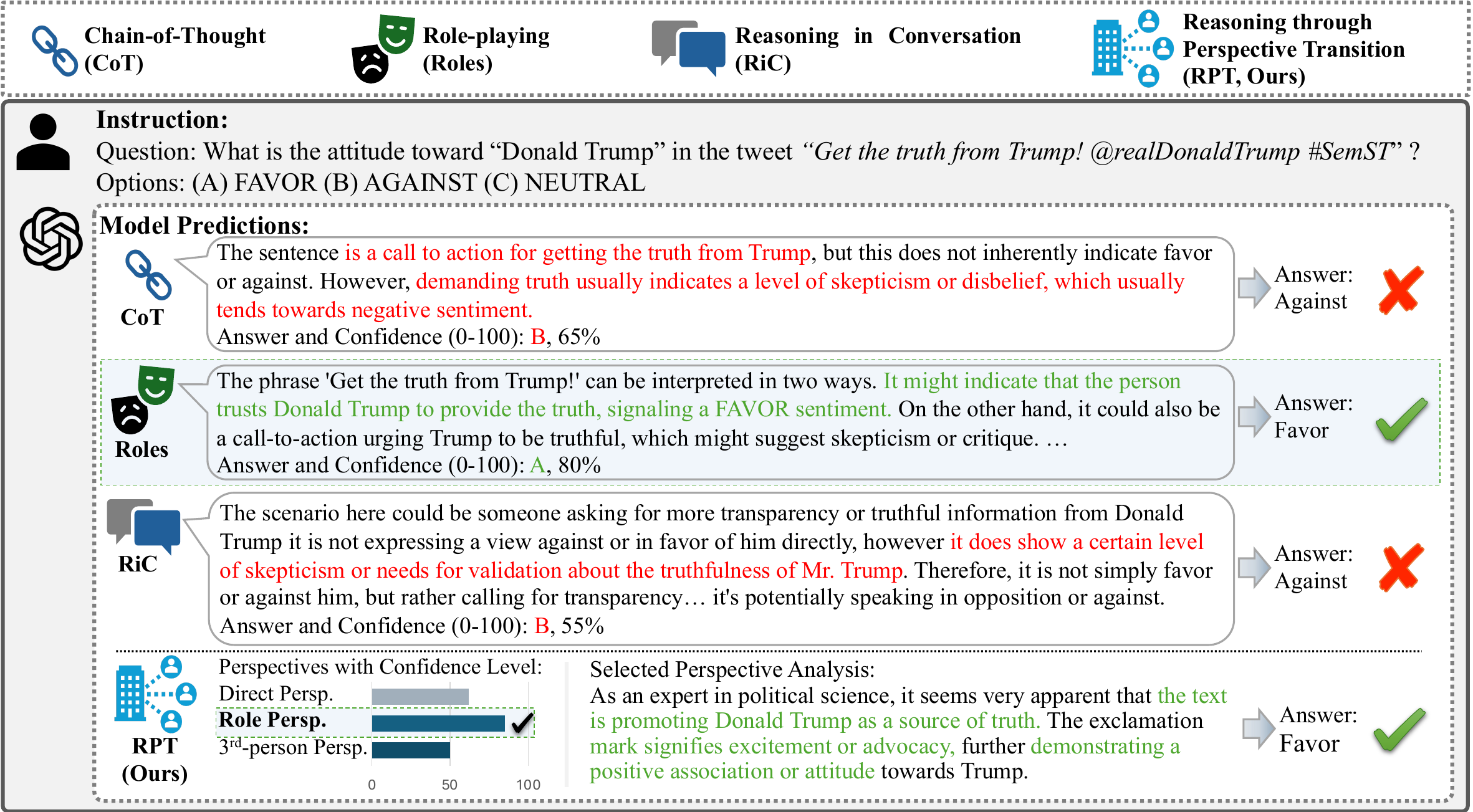}
    \caption{Case of SemEval task. We use GPT-4 to analyze attitudes toward Donald Trump. Our RPT method effectively guides the model in selecting appropriate perspectives for stance detection.
    } \label{fig:case_trump}
\end{figure*}

\subsection{Analysis on the Number of Shots}
\label{sec:number_of_shots}
As shown in Figure~\ref{fig:fig4}, we specify the number of shots and study the performance difference compared to the original RPT on SemEval. We observe that performance is lower when fewer shots are selected, as the model is unfamiliar with the task and method. As the number of shots increases, performance improves. However, in some circumstances when the number of shots reaches three or more, performance declines.

On one hand, LLMs exhibit greater flexibility when autonomously evaluating confidence and planning the number of shots during reasoning, allowing them to adapt to unique subjective tasks. On the other hand, providing too many examples may lead to increasing the inference cost, raising the risk of over-fitting, and challenging the instruction-following ability of LLMs.  Overall, under the majority of settings for each dataset, RPT achieves the best performance, demonstrating its generalization ability and versatility (See Appendix~\ref{app:full_shots} for full results and analysis). 

\subsection{Case Study}

In Figure~\ref{fig:case_trump}, we showcase an example from the SemEval stance detection dataset to highlight the effectiveness of the RPT method in subjective reasoning tasks. Unlike baselines such as CoT and Role-playing, which sometimes emphasize skepticism or negative sentiment without fully accounting for context, RPT evaluates multiple perspectives, including direct and third-person analyses. For example, CoT and RiC interpret the phrase “Get the truth from Trump!” as reflecting skepticism or disbelief, leading to an “AGAINST” prediction. In contrast, RPT dynamically selects the most confident perspective, reasoning that the exclamation mark and phrase suggest advocacy or favor toward Trump. This ability to transition between and rank perspectives makes RPT more adaptable and effective in subjective reasoning tasks compared to single-perspective baselines (See Appendix~\ref{app:cases} for more cases).

\subsection{Analysis on Keyword Statistics}
\begin{figure}[t!]
    \centering
        \includegraphics[width=0.48\textwidth]{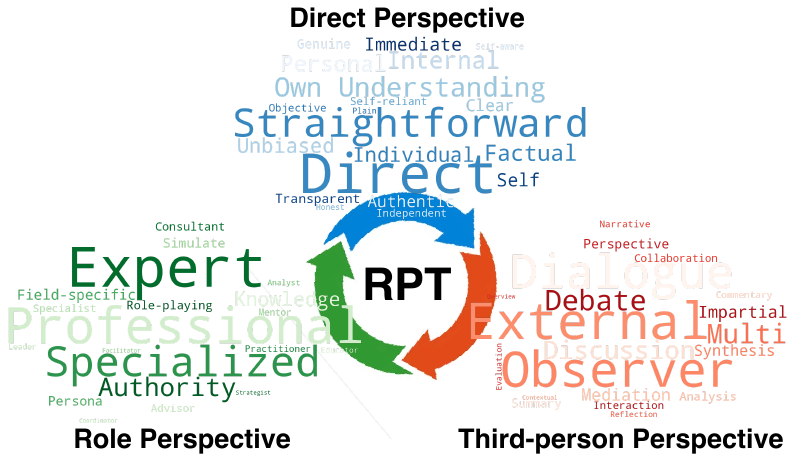}
        \caption{Keyword statistics of different perspectives in the RPT pipeline.}
        \label{fig:wc}
\end{figure}
As shown in Figure~\ref{fig:wc}, to further investigate the characteristics of different perspectives in the RPT pipeline during reasoning, we conduct a keyword frequency analysis for the three perspectives in the RPT pipeline. After removing stopwords and irrelevant prompt words, we can observe the following reasoning characteristics for each perspective: the direct perspective tends to perform straightforward reasoning; the role perspective leans towards adopting different expert roles and contexts; and the third perspective excels in discussions and dialogues. The uniqueness of each perspective underscores the necessity of RPT's dynamic perspective selection.
\section{Conclusion}

In this paper, we introduce RPT,  a novel method that achieves multi-perspective reasoning and integration by exploring diverse perspectives and ranking them based on confidence. Comprehensive experiments conducted on GPT-4, GPT-3.5, Llama-3, and Qwen-2 demonstrate that RPT effectively integrates various perspectives, enhancing the subjective task-solving capabilities of LLMs without significantly increasing inference costs. This work highlights how LLMs can better handle the fluidity of subjective reasoning, even in the absence of nuanced understanding of perspectives or personal biases. Future research directions include incorporating more reasoning perspectives, refining adaptive perspective taxonomies, and expanding applications.

\section*{Limitations}
First, in designing the RPT pipeline, we categorize perspectives into three types based on related works. Although RPT and many inference paradigms involved in the baselines are orthogonal and combinable, this taxonomy could still be further refined, for example, by adopting alternative categorization methods or employing a more fine-grained division. 
Second, RPT directly selects perspectives rather than methods. We consider perspectives as a meta-method, meaning that RPT can be combined with other methods to achieve better performance. 
Thirdly, RPT operates within a single round of dialogue, without accounting for multi-turn conversations or result feedback. In the future, exploring multi-turn dialogue or multi-agent perspective writing could be a promising direction.

\section*{Ethics Statement}
This paper uses widely available datasets, including stance detection, sarcasm detection, and cultural comparison, along with LLM-generated responses, solely to validate the proposed method without reflecting any stance or bias from the authors.


\bibliography{anthology, custom}
\bibliographystyle{acl_natbib}

\appendix

\section{Details of Ablation Study}
\label{app:detail}

As shown in Table~\ref{tab:ablation_detail}, we present the complete and detailed results of the ablation experiments. By removing one, two, and all three perspectives, we demonstrate the effectiveness of RPT. 
Based on the number of perspectives removed, we divide the ablation experiments into three groups. It can be seen that all the perspectives involved in RPT are beneficial. Meanwhile, restricting the range of perspective selection also results in performance degradation.

\begin{table}[htbp]
    
\small
    \centering
    \resizebox{1.0\linewidth}{!}{
    \begin{tabular}{lrrrrr}
    \toprule 
           \textbf{Method}& \bf{SemEval}&\textbf{SocNorm} & \textbf{e-SocNorm} & \textbf{CALI} & \textbf{AVG.} \\
         \midrule
         \textbf{RPT (Ours)}  &$\bf{81.43}$&$\bf{59.81}$&$\bf{77.57}$&$\bf{61.13}$ &$\bf{69.99}$ \\
             \addlinespace[1pt]
    \cdashline{1-6}
    \addlinespace[3pt]
    \multicolumn{6}{c}{ (\textit{removing 1 perspective})} \\
         \ \textit{w/o} DP &{$\downarrow0.83$} &{$\downarrow1.85$}&{$\downarrow1.44$}&{$\downarrow1.14$}&  {$\downarrow1.32$}\\
         \ \textit{w/o} RP&{$\downarrow1.24$} &{$\downarrow3.35$}&{$\downarrow1.71$}&{$\downarrow1.82$}&{$\downarrow2.03$} \\
         \ \textit{w/o} TP&{$\downarrow1.44$} &{$\downarrow3.75$}&{$\downarrow1.88$}&{$\downarrow3.05$}&{$\downarrow2.53$} \\

         \addlinespace[1pt]
    \cdashline{1-6}
    \addlinespace[3pt]
    \multicolumn{6}{c}{ (\textit{removing 2 perspectives})} \\
         \ \textit{w/o} DP\&RP &{$\downarrow2.14$} &{ $\downarrow8.41$}&{$\downarrow5.05$}&{$\downarrow5.00$}&{$\downarrow5.15$} \\
         \ \textit{w/o} DP\&TP &{$\downarrow2.04$} &{ $\downarrow7.35$}&{$\downarrow7.88$}&{$\downarrow5.14$}&{$\downarrow5.60$} \\
         \ \textit{w/o} RP\&TP &{$\downarrow2.78$} &{ $\downarrow8.82$}&{$\downarrow9.27$}&{$\downarrow5.05$}&{$\downarrow6.48$} \\
         \addlinespace[1pt]
    \cdashline{1-6}
    \addlinespace[3pt]
    \multicolumn{6}{c}{ (\textit{removing 3 perspectives})} \\
         \ \textit{w/o} DP\&RP\&TP &{$\downarrow3.93$} &{ $\downarrow9.36$}&{$\downarrow9.76$}&{$\downarrow7.37$}&{$\downarrow7.60$} \\
         
        \bottomrule
    \end{tabular}}
         \caption{Detailed results of ablation study of our proposed RPT method with GPT-3.5 in zero-shot settings. DP: Direct perspective. RP: Role Perspective. TP: Third-Person Perspectives.}
         \label{tab:ablation_detail}
\end{table}

\begin{table*}[htbp]
	\centering
    \scalebox{0.65}{
	\begin{tabular}{clccccccccccccc}
 \toprule
        \multirow{3.5}*{\bf{Type}}&\multirow{3.5}*{\bf{Method}}&\multicolumn{3}{c}{\bf{Linguistic Rhetoric}}&\multicolumn{2}{c}{\bf{Disambiguation QA}}&\multicolumn{2}{c}{\bf{Stance Detection}}&\multicolumn{3}{c}{\bf{Cultural-Related}}&\multicolumn{2}{c}{\bf{Traditional NLI}} &\multirow{3.5}*{\bf{\textsc{Avg.}}}\\
        \cmidrule(lr){3-5}\cmidrule(lr){6-7}\cmidrule(lr){8-9}\cmidrule(lr){10-12}\cmidrule(lr){13-14}
        &&\textbf{Metaphor}&\textbf{SNARKS}&\textbf{Humor}&\textbf{Pronoun}&\textbf{Anach.}&\textbf{SEQ}&\textbf{SemEval}&\textbf{SocNorm}&\textbf{e-SocNorm}&\textbf{CALI}&\textbf{Entail.}&\textbf{IPA}\\
        &&(Acc.)&(Acc.)&(Acc.)&(Acc.)&(Acc.)&(Acc.)&(F1)&(F1)&(F1)&(Acc.)&(Acc.)&(Acc.)&\\
    \midrule
    
D&\textbf{RPT (Ours)}&$\bf{91.76} $ & $\bf{87.29} $ & $\bf{70.00} $ & $\bf{65.12} $ & $\bf{73.48} $ & $\bf{99.13} $ & $\bf{81.43} $ & $\bf{59.81} $ & $\bf{77.57} $ & $\bf{61.13} $ & $\bf{88.57} $ & $\bf{80.00} $ & $\bf{77.94}$\\
 \addlinespace[1pt]
    \cdashline{1-15}
    \addlinespace[3pt]
    D&\textit{RPT (second)}&$83.82 $ & $81.77 $ & $56.59 $ & $56.20 $ & $72.61 $ & $98.26 $ & $80.61 $ & $43.52 $ & $61.53$ & $51.36 $ & $77.14 $ & $64.29 $ & $68.98$\\
D&\textit{RPT (lowest)}&$79.12 $ & $79.01 $ & $37.98 $ & $38.76 $ & $69.13 $ & $96.52 $ & $78.89$ & $38.03 $ & $51.54 $ & $48.64 $ & $57.14 $ & $56.35 $ & $60.93$\\
D&\textit{RPT (random)}&$85.88$ & $80.11 $ & $53.49 $ & $58.91 $ & $70.43 $ & $97.39$ & $80.56 $ & $45.15 $ & $67.99 $ & $54.09 $ & $67.14 $ & $67.46 $ & $69.05$\\
     \bottomrule
    \end{tabular} }
  \caption{Analysis on confidence-based perspective selection. \textit{RPT (random)} represents the result of random prediction with uniform probability,  \textit{RPT (second)} represents selecting the perspective with the second highest confidence, and \textit{RPT (lowest)} means choosing the most unconfident perspective.}
	\label{table:mainresults4}
    \vspace{-1em}
\end{table*}

\section{More Analysis}
\label{app:more_detail}

RPT ranks different perspectives based on confidence levels without relying on external information. In this section, using the zero-shot GPT-3.5 experiment as an example, we force the LLM to select the perspective with the second highest confidence, the lowest confidence, and a randomly chosen perspective during RPT inference. 

As shown in Table~\ref{table:mainresults4}, the lower the confidence of the selected perspective, the poorer the performance of LLM. When randomly selecting perspectives, the performance of the LLM is also worse than that of the perspective with the highest confidence. This shows that ranked perspectives based on confidence levels are effective, explaining the underlying mechanism by which RPT improves performance.

\subsection{Analysis on the Correlation between Confidence and Accuracy}
\begin{figure}[htbp!]
    \centering
     
 \includegraphics[width=0.5\textwidth]{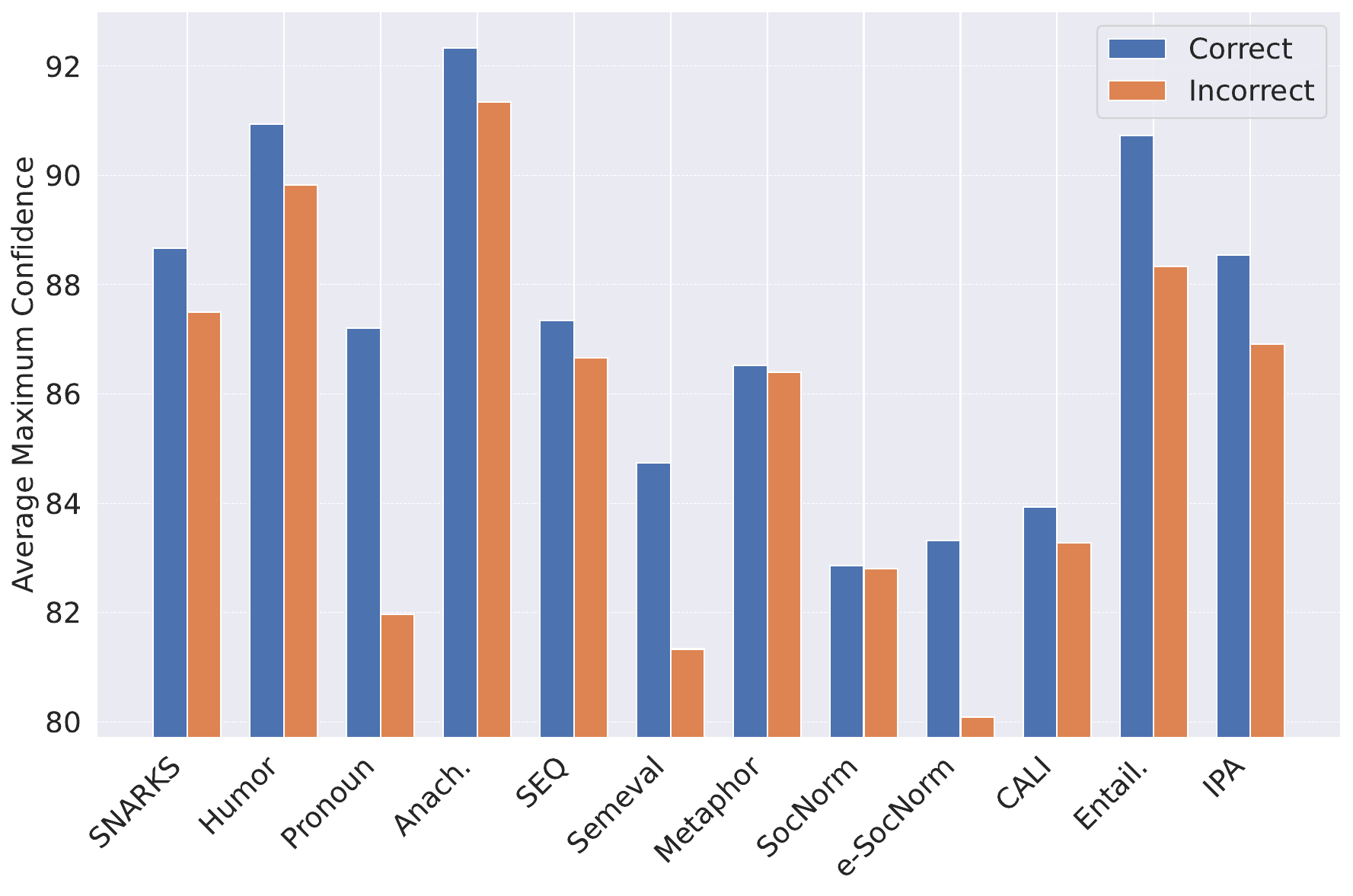}
    \caption{Analysis on the correlation between confidence and accuracy.} \label{fig:confidence}
\end{figure}
In RPT, we use LLM itself to judge the confidence of perspectives for a given input, allowing the model to rank and switch among perspectives accordingly. As shown in Figure~\ref{fig:confidence}, using the GPT-3.5 model as an example, we analyze the relationship between predicted confidence and the actual accuracy. We can observe that when confidence exceeds the threshold of approximately 70\%, the accuracy of the chosen perspective is significantly higher. This indicates that LLMs are capable of ranking the confidence of perspective for a specific input based on confidence levels.

     

\begin{figure}[t!]
    \centering
    \includegraphics[width=\columnwidth]{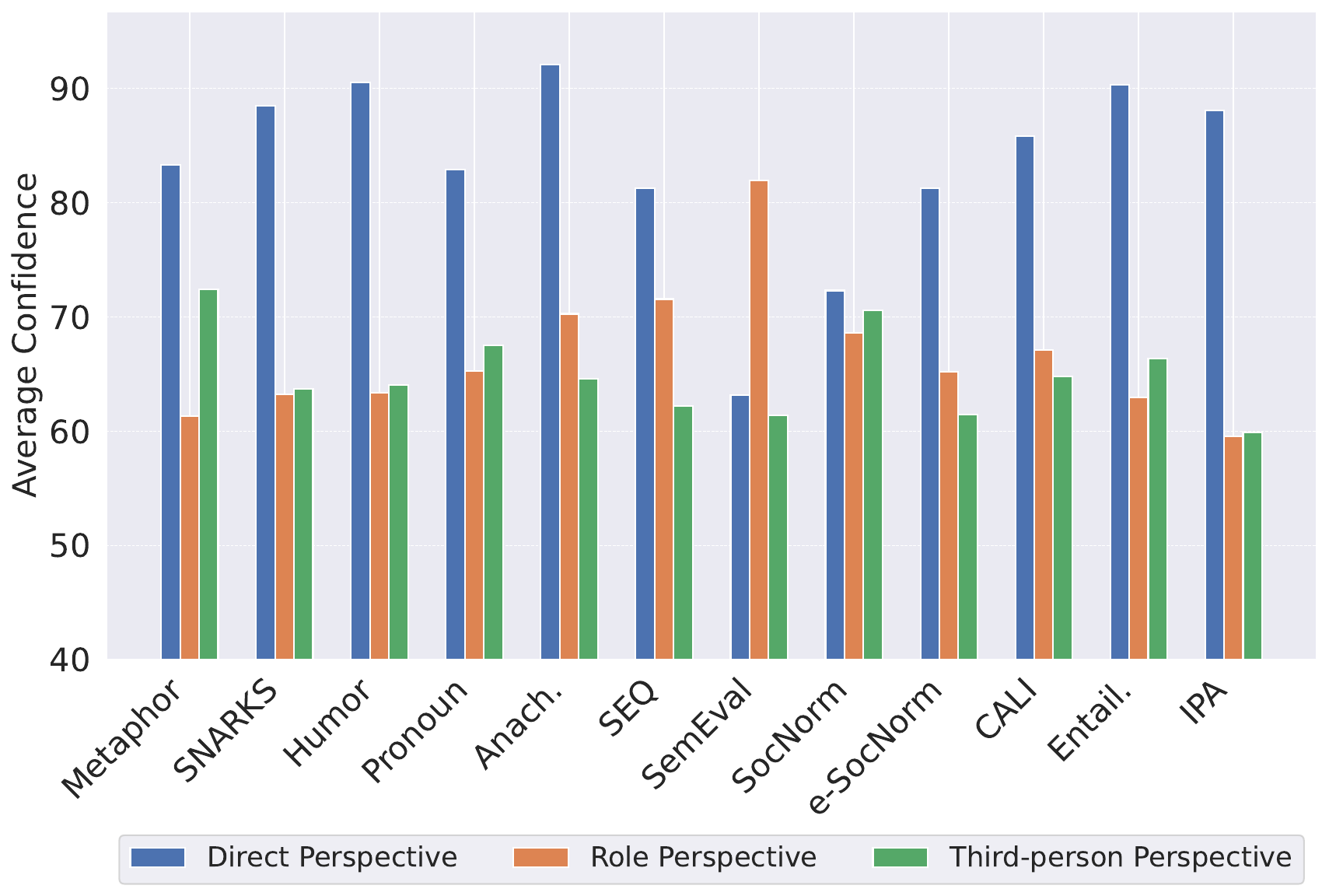}
    \caption{The averaged confidence level by our RPT method in different datasets.}
    \label{fig:turns_keywords}
    \vspace{-1em}
\end{figure}

\begin{figure}[t!]
    \centering
    \includegraphics[width=\columnwidth]{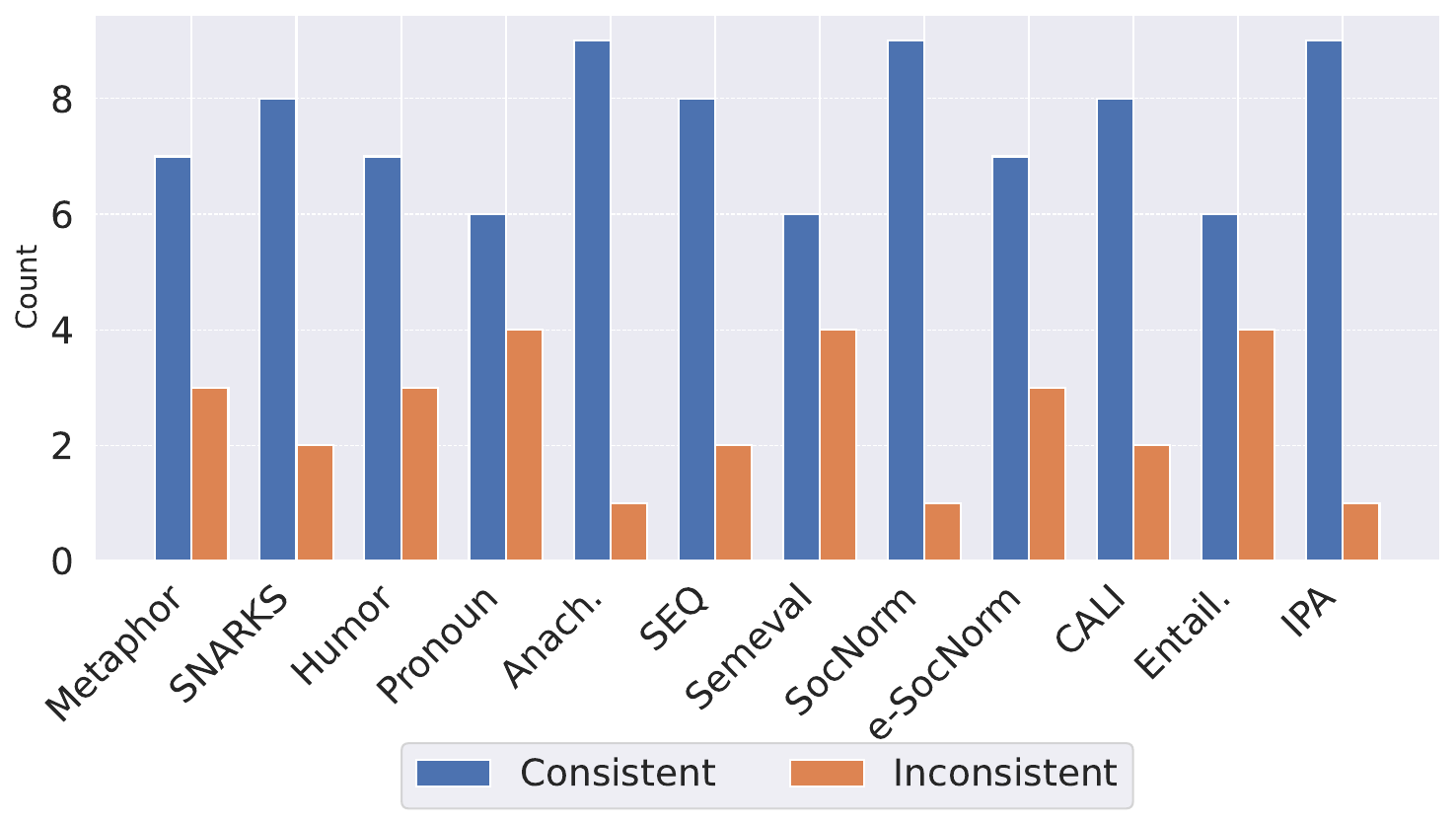}
    \caption{Human evaluation consistency on confidence.}
    \label{tab:confidence_stat}
\end{figure}

Using GPT-3.5 as an example, we report in Figure~\ref{fig:turns_keywords} the average confidence for each dataset. Figure~\ref{tab:confidence_stat} shows the human evaluation consistency when estimating the confidence.  We find that when evaluating confidence, the estimation of the LLM are highly correlated with those of human experts, indicating that the LLM has the ability to evaluate confidence and select perspectives. 

\begin{figure}[htbp!]
    \centering
 \includegraphics[width=0.48\textwidth]{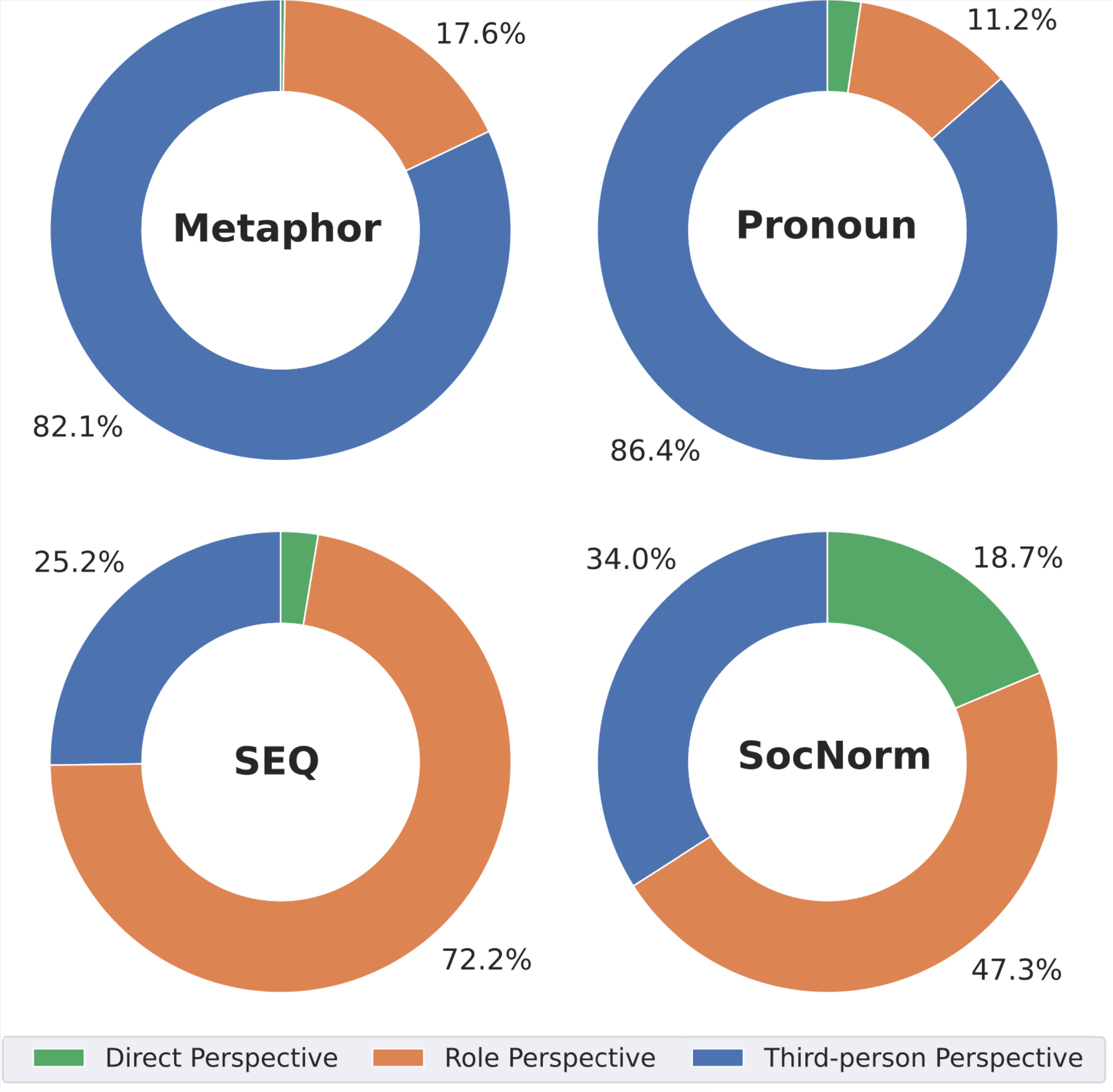}
    \caption{The proportion of selected perspectives for different input questions in each dataset.} \label{fig:pie}
\end{figure}

Moreover, RPT generally performs better within high-confidence perspectives, indicating that confidence-based perspective ranking is efficient when choosing among perspectives. In Figure~\ref{fig:pie}, we present the proportion of different perspectives used on each dataset, showing that different datasets have different perspective biases. This suggests that, compared to a single perspective, PRT offers perspective flexibility, which helps RPT achieve optimal performance. 

\subsection{More Cases of RPT}
\label{app:cases}
\begin{figure*}[htbp]
    \centering
    \includegraphics[width=0.96\textwidth]{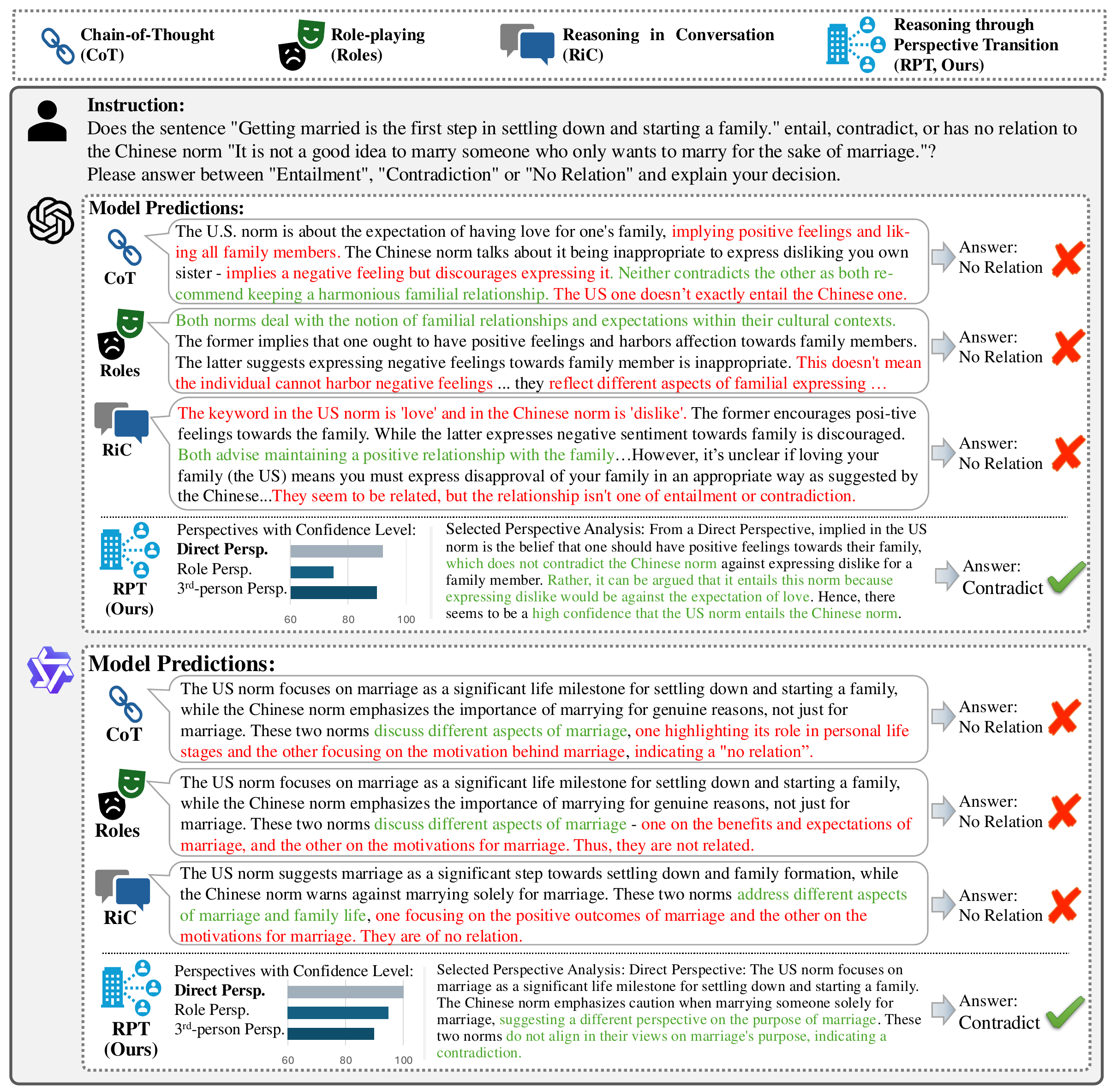}
    \caption{Cases of SocNorm task. We provide results of two widely used models, GPT-4 and Qwen-2. In each of the cases, our proposed method successfully spots the actual meaning behind these norms and leads to the correct answer.} \label{fig:case_socnli}
\end{figure*}

\begin{figure*}[htbp]
    \centering
    \includegraphics[width=0.9\textwidth]{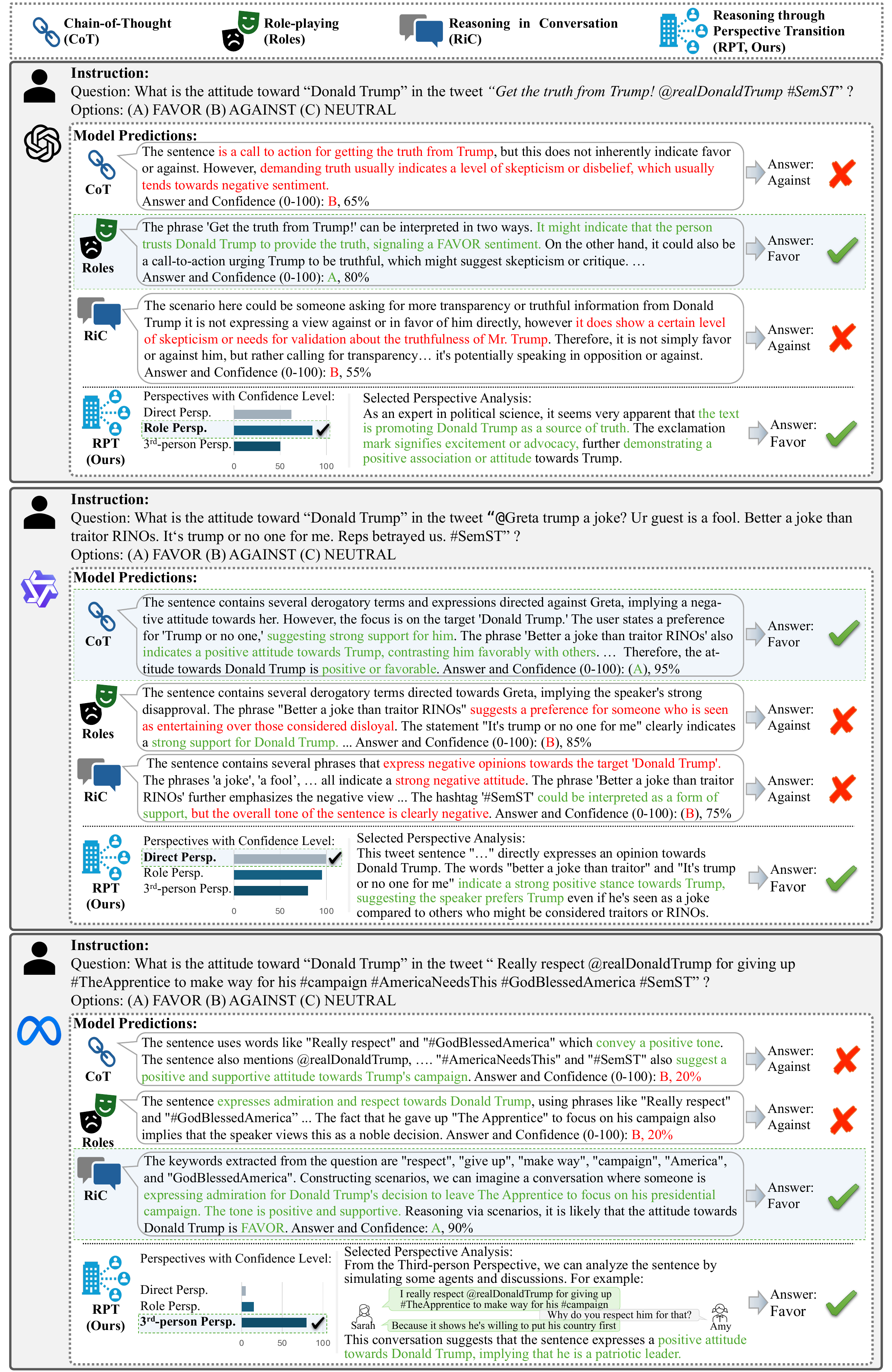}
    \caption{Cases of SemEval task. We provide detailed responses of three models, GPT-4, Qwen-2, and Llama-3, regarding the attitude towards Donald Trump. Our method prompts models to successfully selects suitable perspectives to solve stance detection problems.} \label{fig:case_stance}
\end{figure*}

In Figure~\ref{fig:case_stance} and Figure~\ref{fig:case_socnli}, we present several examples from a culturally related NLI task SocNorm and a stance detection dataset SemEval. Take the second case in SemEval task as an example, baseline methods captures strong negative emotions in the input through words like ``joke'', ``fool'' and ``betray'' and makes judgments about the speaker's attitude toward Trump based on these cues, overlooking the potential underlying implications of the text. However, RPT evaluates and selectes the third-person perspective, providing the correct analysis by simulating some agents and discussions, illustrating the effectiveness of the RPT in subjective reasoning tasks.

\begin{figure*}[h]
    \centering
    \begin{subfigure}[b]{0.48\textwidth}
        \includegraphics[width=\textwidth]{imgs/semeval-shots.pdf}
        \caption{SemEval}
        \label{fig:image1}
    \end{subfigure}
    \hfill
    \begin{subfigure}[b]{0.48\textwidth}
        \includegraphics[width=\textwidth]{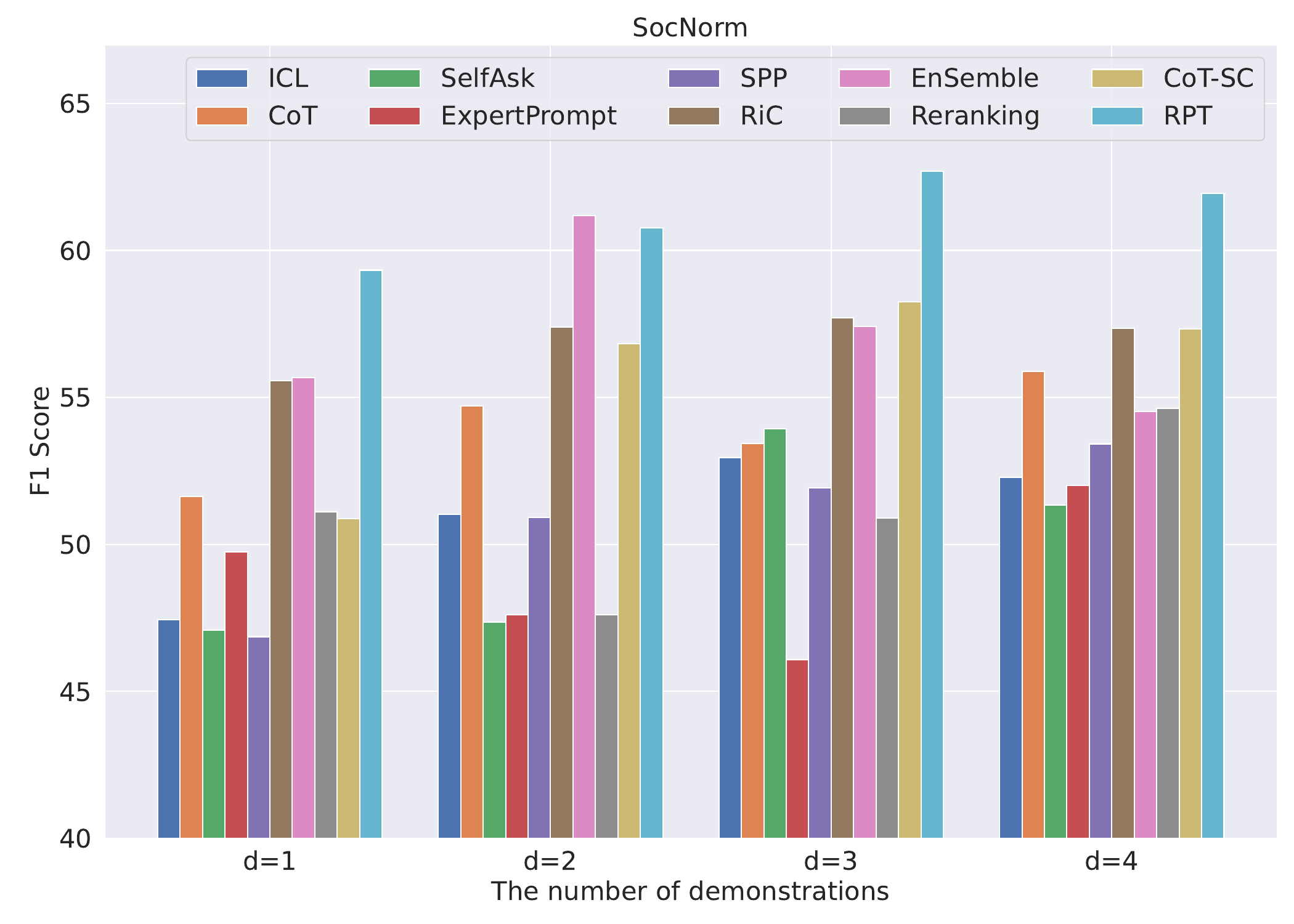}
        \caption{SocNorm}
        \label{fig:image2}
    \end{subfigure}
    
    \vspace{1em} 
    
    \begin{subfigure}[b]{0.48\textwidth}
        \includegraphics[width=\textwidth]{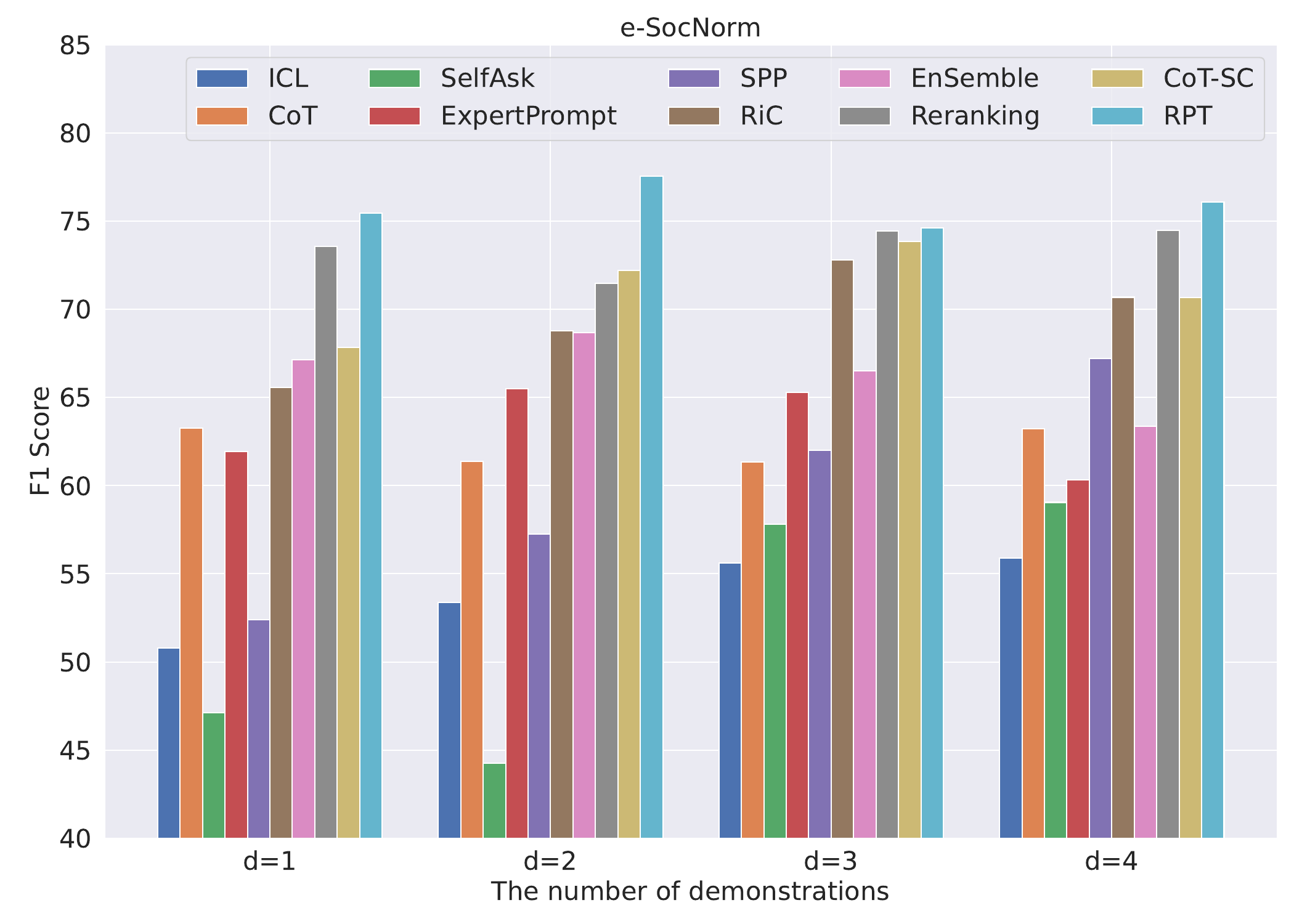}
        \caption{e-SocNorm}
        \label{fig:image3}
    \end{subfigure}
    \hfill
    \begin{subfigure}[b]{0.48\textwidth}
        \includegraphics[width=\textwidth]{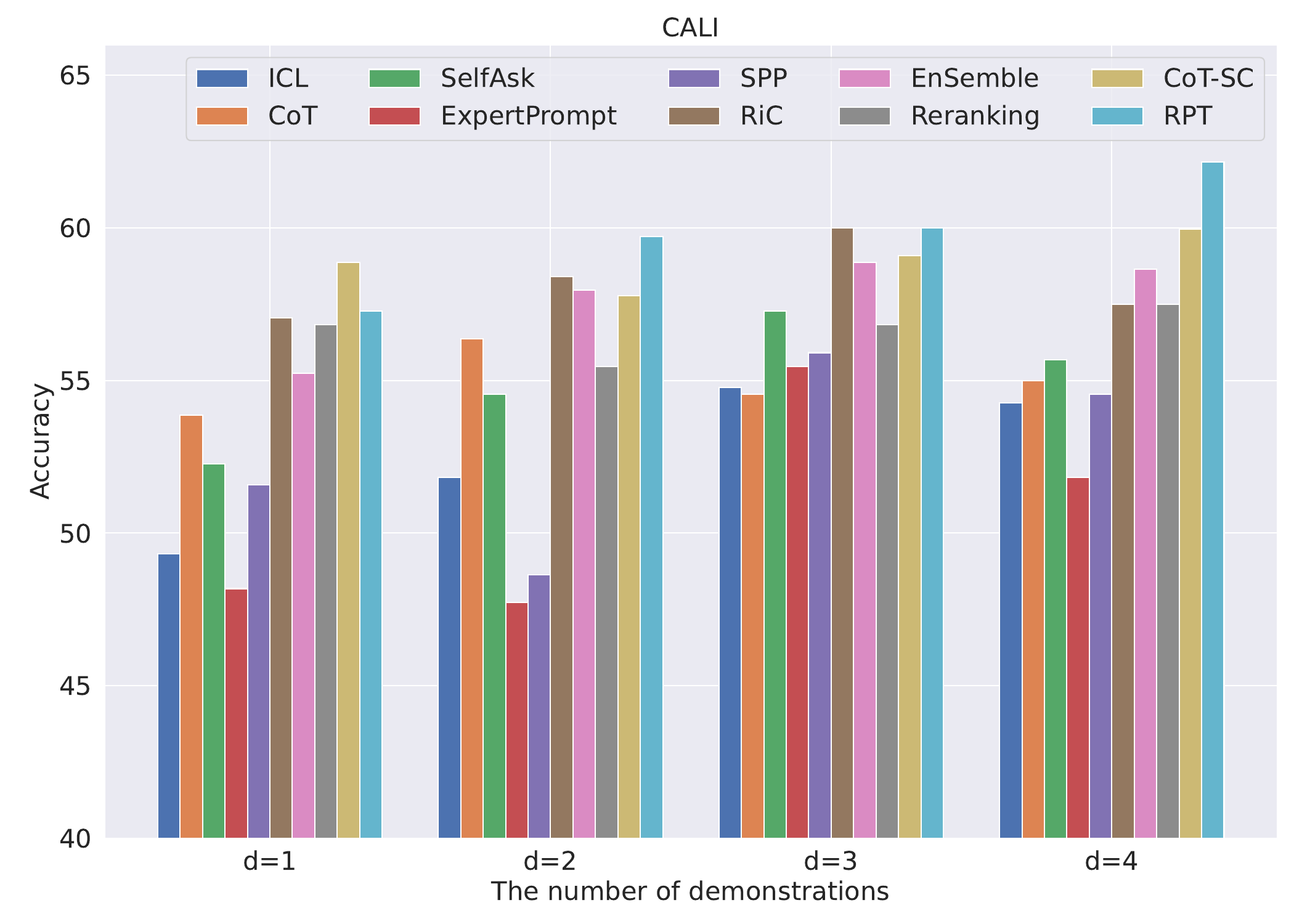}
        \caption{CALI}
        \label{fig:image4}
    \end{subfigure}
    
    \caption{The performance of baselines and our RPT method by using different numbers of demonstrations ($d=1,2,3,4$) in few-shot settings.}
    \label{fig:fig4v}

\end{figure*}

\subsection{Comparison with Ensemble-based Methods}
\label{sec:compare_ensemble}
Instead of oversampling responses and simply making a selection using model confidence, 
the core idea of RPT is that LLMs can leverage their internal priors and the given prompt to directly select the perspective with the highest posterior accuracy. Unlike conventional ensembling approaches, RPT assigns a single optimal perspective to each query without attempting inferencing on multiple perspectives per instance before making a decision. This means that:

\begin{itemize}
    \item RPT does not ensemble multiple responses.
\item RPT does not significantly increase the generation window like ensembling does.
\item RPT does not rely on oversampling responses before selection.
\end{itemize}

Instead, RPT possesses the following features:

\begin{itemize}
    \item RPT selects a single perspective after a brief internal deliberation based on the input prompt and then conducts inference solely from that perspective.

\item The computational cost of RPT is comparable to using a single perspective.

\item Despite its simplicity, RPT achieves performance comparable to or better than ensemble methods.
\end{itemize}

For example, given $k$ candidate perspectives, an ensemble method incurs a response length/inference cost of $k \times $
 cost, with a performance of p. In contrast, RPT has a response length/inference cost of approximately $1 \times $ 
 cost, while achieving performance $ \ge
 p$. We believe that LLMs' internal priors can be leveraged to preemptively plan and evaluate reasoning chains across different perspectives, allowing them to anticipate ensemble outcomes before actually performing ensembling, thus eliminating unnecessary computational overhead. This potential has not been explored before. Analogous to Mixture of Experts (MoE), RPT can be regarded as a form of ``Mixture of Perspectives'', where exactly one perspective is activated per inference. Thus, we believe that RPT is a highly novel approach.

Therefore, instead of simply selecting perspective, we propose the dynamic process of shifting between different perspectives when handling various tasks, rather than merely selecting a single perspective. It highlights the following two aspects: 
\begin{itemize}
    \item Dynamism: The model autonomously transitions from one perspective to another, rather than making a static selection. This process is an inherent decision made by the model itself. \item Variability: In practice, this transition occurs through mechanisms such as Chain-of-Thought (CoT), role-playing, multi-agent discussions, and other discourse-based approaches, facilitating perspective shifts. Compared to perspective selection, we emphasize the dynamic and evolving nature of this process.
\end{itemize}
\subsection{Details on Shot Numbers}
\label{sec:shot_number_detail}
In Section~\ref{sec:number_of_shots}, we provide experimental results across multiple shot settings, demonstrating that the 3-shot setting yields the best performance. Moreover, the datasets used in this study primarily consist of binary or ternary classification tasks. Conventionally, 3-shot is a common setting for practical inference. Additionally, both the baseline and our proposed method achieve the best overall performance under this setting, making it a reasonable choice for our primary study on inference cost. Therefore, we choose 3-shot setting to conduct analysis on performance-inference cost experiments.

\subsection{Estimation of Inference Cost}

The purpose of Section~\ref{sec:cost} is to examine the trade-off between inference cost and performance when using different methods with LLMs. Counting the number of output tokens is the most common approach for estimating inference time, as LLMs process input tokens in parallel with relatively low overhead, whereas output generation is auto-regressive and thus dominates inference latency and cost. For common LLMs, the cost of input tokens is lower than output tokens. Including both input and output tokens in the performance-cost scatter plot would introduce bias and deviate from standard practices in the LLM community.

Therefore, we chose to plot a two-dimensional scatter plot of output token count versus performance. Introducing a third variable would unnecessarily complicate the experiment and make it more difficult to draw clear conclusions.

\subsection{More Examples on Subjective Tasks}
\label{sec:more_cases}

Perspective in language is essentially a matter of deixis, how entities are positioned relative to the speaker (user), the addressee (LLM), and a potential third party.
\begin{itemize}
    \item First-person refers directly to the speaker's own viewpoint, emphasizing subjectivity and experiential immediacy. It forms the core of self-reference and establishes speaker agency.  This is the most common LLM reasoning mode, as it activates the model's cognitive framing, directly connecting it to the question and reasoning process.
    \item Second-person focuses on the listener in an interaction, creating direct engagement. It reduces the narrative distance through dialogic framing, commonly used in imperative or persuasive contexts.  
  LLMs can leverage second-person narration to enhance reasoning in ambiguous scenarios by simulating conversational exchange.
  \item Third-person provides an external, often omniscient view of the situation or entities involved.  
  In LLM reasoning, this resembles a form of distancing effect, offering broader perspective through heterodiegetic narration.
\end{itemize}

As shown in the experiment results, our experiments reveal that for subjective tasks, Zero-shot-CoT does not exhibit a significant advantage over Direct Prompting on GPT-3.5 and GPT-4-0613, and even performs slightly worse on Qwen2-7B-Instruct (Direct Prompt: 62.70, Zero-shot-CoT: 61.42), as detailed in Table~\ref{table:mainresults}. This suggests that unlike objective tasks, subjective tasks may not allow for a universally applicable reasoning pathway. This observation aligns with findings by~\citet{wang-etal-2024-reasoning}, which systematically studies the performance of LLMs on subjective tasks. 

\textbf{Example 1:}
\begin{itemize}
    \item Question: Identify whether the sentence ``her smile was a blinding sun'' is a paraphrase of the metaphoric sentence ``her smile was wonderful and fulfilling to see''.
    \item Chain-of-Thought Reasoning: While both sentences express admiration for her smile, the second sentence lacks the vivid imagery and specific emotional intensity that the first sentence provides. Thus, the second sentence is not an accurate paraphrase of the first. The answer is False.
\end{itemize}

\textbf{Example 2}:
\begin{itemize}

\item Question: Identify whether the following examples are intended to be a joke (with dark humor) or not ``I have a fish that can breakdance! Only for 20 seconds though, and only once''.
\item Chain-of-Thought Reasoning: The phrase is light-hearted and whimsical about a fish that can breakdance, suggesting a humorous intent rather than a dark theme. Therefore, it does not seem to fit the category of dark humor. The answer is Not.
\end{itemize}

\textbf{Example 3}:
\begin{itemize}
    \item Question: The sentence ``Joseph is very kind'' is a paraphrase of the metaphoric sentence ``Joseph has the heart of a lion''.
    \item RiC Reasoning: The second sentence suggests that Joseph possesses qualities associated with a lion, while the first sentence only mentions kindness. The answer is False.
\end{itemize}
\subsection{Discussion on Overconfidence and Underconfidence}

Regarding the inherent overconfidence and underconfidence phenomena in LLMs, previous research~\cite{eot} and our experiments (as shown in Figure~\ref{fig:ablation}) indicate that the higher the confidence of an LLM’s response, the more likely it is to be correct. We have further investigated the impact of different confidence levels on data selection (as shown in Table~\ref{table:mainresults4}) by explicitly providing confidence scores for different perspectives and selecting perspectives based on these confidence levels.

Our experiments also show that automatically selecting the perspective with the highest confidence achieves an average performance of 77.94, significantly outperforming other baseline methods. In contrast, selecting the second-highest confidence perspective results in an average performance of 68.98, which is similar to random selection (69.05). The lowest-confidence selection, however, performs notably worse, with an average performance of only 60.93, indicating that choosing perspectives with the lowest confidence leads to poor outcomes.

In Figure~\ref{tab:confidence_stat}, by comparing human evaluation scores with LLM-generated confidence scores after normalization, we find that LLMs' confidence predictions are generally accurate. As shown in Figure~\ref{fig:confidence}, a per-instance comparison demonstrates a high correlation between LLM-predicted confidence and actual accuracy. In future research, providing additional supervision signals to models based on dataset characteristics and input prompt designs may further mitigate overconfidence and underconfidence issues.

\subsection{Full Results of Analysis on the Number of Shots}
\label{app:full_shots}
The full results of analysis on the number of shots is shown in Figure~\ref{fig:fig4v}. The results are consistent across the four datasets. It can also be observed that compared to zero-shot settings, few-shot settings are not always helpful, which is consistent with the findings by~\citet{deepseek2025r1}. The following factors may contribute to the observed trends:

\begin{itemize}
\item As shown in Figure~\ref{fig:fig4}, increasing the number of shots does not always enhance LLM performance. While providing a few examples can help the model handle unfamiliar tasks and follow instructions, excessive shots can lead to diminishing returns or performance degradation. We chose the 3-shot setting because most datasets consist of binary or ternary classification tasks.
\item The length of the input prompt can impact LLM inference. Excessive in-context examples may distract the model, reducing its ability to follow instructions precisely, which can negatively impact performance.
\item The added examples may introduce noise, affecting reasoning performance. For instance, gpt-3.5-turbo-1106 tends to overfit the provided cases, often favoring the answer from the last example presented in the prompt.
\item The inherent openness and uncertainty of subjective tasks may contribute to this phenomenon. Subjective task responses are often ambiguous, and in few-shot settings, methods like Ensemble, Reranking, and CoT-SC may be influenced by the provided examples, potentially introducing biases and limiting generalization. In contrast, zero-shot performance relies on pre-trained knowledge, making it more stable. Ensemble methods are susceptible to errors from individual models, reranking depends on the quality of initial generations, and CoT-SC may reduce reasoning diversity, diminishing its advantage in maintaining consistency.
\end{itemize}

\subsection{Using Small Models as Classifiers}
We experiment with using small discriminative models, such as bert-base-uncased, as classifiers. However, due to the complexity of the datasets involved in RPT and the lack of distinctive content features, the classification performance was generally poor. Specifically, we use a large model to run each of the different perspective methods on the training set. For samples where the answer was correct, we labeled them according to a priority order: 1 (Direct), 2 (Role-Play), 3 (Third-Person), and then split the data into training and test sets with an 8:2 ratio. The results are shown in Table~\ref{tab:small_model_hyperparam}, with the highest accuracy reaching only 48\%.

\begin{table}[t]
\small
    \centering
    \resizebox{0.8\linewidth}{!}{
    \begin{tabular}{lcc}
    \toprule
    \textbf{Hyperparameters} & \textbf{Accuracy (\%)} & \textbf{F1-Score (\%)} \\
    \midrule
    epoch: 3, lr: 2e-5 & 36 & 23 \\
    epoch: 4, lr: 2e-5 & 33 & 32 \\
    epoch: 4, lr: 1e-5 & 30 & 28 \\
    epoch: 4, lr: 2e-4 & 47 & 35 \\
    epoch: 4, lr: 5e-4 & \underline{48} & 31 \\
    epoch: 8, lr: 2e-5 & 46 & \underline{35} \\
    \bottomrule
    \end{tabular}
    }
    \caption{Performance of small model classifier with different training hyperparameters.}
    \label{tab:small_model_hyperparam}
\end{table}

These results suggest that using small models is not a feasible approach. Moreover, introducing additional parameters for perspective ranking increases inference overhead. In contrast, RPT leverages the LLM’s own capabilities to assess the suitability of different perspectives on a given input without requiring additional training or parameters, achieving performance gains more efficiently.
\end{document}